\pdfoutput=1

\documentclass[11pt]{article}
\PassOptionsToPackage{table,dvipsnames}{xcolor}

\usepackage[preprint]{acl}

\usepackage{algorithm}
\usepackage{algpseudocode} 
\usepackage{times}

\usepackage[T1]{fontenc}
\usepackage[utf8]{inputenc}
\usepackage{microtype}
\usepackage{inconsolata}

\usepackage{svg}

\usepackage{graphicx}
\usepackage{amsmath}
\usepackage{listings}
\usepackage{subcaption}
\usepackage{enumitem}
\usepackage{stfloats}

\lstdefinelanguage{json}{
  morestring=[b]",
  morecomment=[l]{//},
  morekeywords={true,false,null},
  sensitive=false,
}

\title{\textit{SurveyGen-I}:  Consistent Scientific Survey Generation with Evolving Plans and Memory-Guided Writing}

\author{
\textbf{Jing Chen}\textsuperscript{1,2}\thanks{~~Equal contribution.}~, 
\textbf{Zhiheng Yang}\textsuperscript{1}\footnotemark[1], 
\textbf{Yixian Shen}\textsuperscript{1}, 
\textbf{Jie Liu}\textsuperscript{1}, \\
\textbf{Adam Belloum}\textsuperscript{1}, 
\textbf{Chrysa Papagianni}\textsuperscript{1}, 
\textbf{Paola Grosso}\textsuperscript{1} \\
\textsuperscript{1}University of Amsterdam, the Netherlands \\
\textsuperscript{2}Vrije Universiteit Amsterdam, the Netherlands \\
\texttt{j.chen12@student.vu.nl} \\
\texttt{\{z.yang, y.shen, j.liu, a.s.z.belloum, c.papagianni, p.grosso\}@uva.nl}
}

\begin{document}
\maketitle
\begin{abstract}

Survey papers play a critical role in scientific communication by consolidating progress across a field. 
Recent advances in Large Language Models (LLMs) offer a promising solution by automating key steps in the survey-generation pipeline, such as retrieval, structuring, and summarization. However, existing LLM-based approaches often struggle with maintaining coherence across long, multi-section surveys and providing comprehensive citation coverage.
To address these limitations,  we introduce SurveyGen-I, an automatic survey generation framework that combines coarse-to-fine retrieval, adaptive planning, and memory-guided generation. 
SurveyGen-I first performs survey-level retrieval to construct the initial outline and writing plan, and then dynamically refines both during generation through a memory mechanism that stores previously written content and terminology, ensuring coherence across subsections.
When the system detects insufficient context, it triggers fine-grained subsection-level retrieval.
During generation, SurveyGen-I leverages this memory mechanism to maintain coherence across subsections. 
Experiments across four scientific domains demonstrate that SurveyGen-I consistently outperforms previous works in content quality, consistency, and citation coverage.
The code is available at \url{https://github.com/SurveyGens/SurveyGen-I}.

\end{abstract}

\section{Introduction}

The exponential expansion of scholarly literature, with thousands of new papers published daily, presents significant challenges for researchers to efficiently acquire and synthesize comprehensive knowledge. Consequently, writing survey papers requires substantial expertise and time commitment from researchers, as it traditionally involves an iterative and labor-intensive process of reading, note-taking, clustering, and synthesis~\citep{carrera2022conduct}. Recent advances in Large Language Models (LLMs) offer a promising solution to this bottleneck by enabling the automation of key steps in the survey-writing pipeline, such as retrieving, organizing, and summarizing large volumes of papers \citep{wang2024autosurvey, liang2025surveyx, yan2025surveyforge, agarwal2024litllm, agarwal2024litllms}.

Despite recent advances, current LLM-based survey generation frameworks remain limited in several key aspects. \textbf{First, literature retrieval scope and depth remain limited.} Most systems rely on embedding-based similarity search over a fixed local paper database~\citep{wang2024autosurvey, yan2025surveyforge}. While efficient, such surface-level matching often fails to identify important papers with different terminology or at a more conceptual level, resulting in incomplete or biased coverage. \textbf{Second, lack of cross-subsection consistency.} Most systems generate all subsections in parallel as isolated units without modeling dependencies across subsections~\citep{wang2024autosurvey, liang2025surveyx, yan2025surveyforge}. This often leads to redundant content, inconsistent terminology, and fragmented discourse. Moreover, they always follow a static, once-for-all outline that cannot adapt to newly generated content, making it difficult to maintain content coherence or integrate emerging insights. \textbf{Finally, indirect citations are often left unresolved.} Retrieval-augmented generation (RAG) typically extracts passages from retrieved papers to support writing. These passages often include indirect citations such as "[23]" and "Smith et al., 2022", which refer to influential prior work not present in the retrieval results. Without tracing these references, the system may miss influential papers, leading to incomplete citation coverage and broken linkage between ideas and their original sources.

To address these limitations, we introduce SurveyGen-I, an end-to-end, modular framework for generating academic surveys with consistent content and comprehensive literature coverage.
\textbf{First}, SurveyGen-I performs literature retrieval at both the survey-level and subsection-level, augmented with citation expansion and LLM-based relevance scoring. This multi-level retrieval strategy substantially enhances literature coverage and topical relevance.
\textbf{Second}, SurveyGen-I introduces \textbf{PlanEvo}, a dynamic planning mechanism powered by an evolving memory that continuously accumulates terminology and content from earlier generated subsections. This memory is used to construct the outline and a dependency-aware writing plan that captures the logical and conceptual relationships between subsections, allowing foundational topics to be generated before more advanced or derivative ones. As writing progresses, both the outline and plan are continually refined based on the updated memory, ensuring consistent terminology and coherent content flow across the survey.
\textbf{Finally}, SurveyGen-I introduces \textbf{CaM-Writing}, which combines a citation-tracing module that detects indirect references in retrieved passages and resolves them back to their original source papers, with memory-guided generation that uses the evolving memory to maintain coherent terminology and content across the survey.

Extensive results highlight the strengths of SurveyGen-I across multiple dimensions of academic survey generation. Compared to the strongest baseline, SurveyGen-I yields an 8.5\% improvement in content quality, a 27\% increase in citation density, and more than twice as many distinct references, while also demonstrating significantly better citation recency. These improvements show the effectiveness of the system in enabling high-quality and consistent survey generation.

Our contributions are summarized as follows: 
\begin{itemize}[itemsep=0pt, topsep=0pt, parsep=0pt, partopsep=0pt]
   \item We propose \textbf{SurveyGen-I}, a novel framework for high-quality, reference-rich, and consistent survey generation.
   \item We design a multi-stage \textbf{Literature Retrieval} pipeline that combines keyword search, citation expansion, and LLM-based filtering to construct relevant and comprehensive paper sets at both survey and subsection levels.

   \item We introduce \textbf{PlanEvo}, a dynamic planning mechanism that constructs and continuously refines the outline and writing plan based on inter-subsection dependencies and evolving memory, ensuring coherent survey generation.

   \item We develop a \textbf{CaM-Writing} pipeline that combines citation tracing and memory-guided generation to improve reference coverage and ensure consistent, well-structured writing.

\end{itemize}

\section{Related work}
\label{sec:related}

\textbf{Component-Oriented and Hybrid Approaches.}
A longstanding approach to assisting literature surveys has been to tackle the problem in stages, where components handle retrieval, structuring, or writing, etc., independently~\citep{susnjak2025automating, lai2024instruct, li2024chatcite}. 
Early systems organized citation sentences through clustering or classification~\citep{nanba2000classification, wang2018citationas}, or employed rule-based content models~\citep{hoang2010towards, hu2014automatic}. These systems often relied on static heuristics or surface-level topic associations, making them difficult to generalize across domains or maintain narrative coherence.

The rise of LLMs brought a wave of hybrid designs that integrated neural summarization with structured control~\citep{zhang2024chain, fok2025facets, an2024vitality}. Template-based generation~\citep{sun2019automatic} and extractive-abstractive hybrids~\citep{shinde2022extractive} introduced more fluent synthesis but retain rigid structures. Meanwhile, RAG-based methods~\citep{lewis2020retrieval, ali2024automated, agarwal2024litllm} enhanced retrieval fidelity~\citep{gao2023retrieval}, and agent-driven systems like the framework proposed by~\citet{brett2025patience}, RAAI~\citep{pozzobon2024implementation} and AutoSurveyGPT~\citep{xiao2023autosurveygpt} broke down the pipeline into retrieval, filtration, and generation stages.
More recent works emphasize pre-writing planning, such as COI-Agent~\citep{li2024chain}, which organizes references into conceptual chains to enhance topic coverage. 

However, these designs remain fundamentally decomposed: content selection and writing are planned in isolation, and their execution often lacks global coordination across stages.

\paragraph{End-to-End Automated Literature Review/Survey Generation.}

With increasing demand for scalability and consistency, end-to-end frameworks have emerged to streamline the full pipeline from retrieval to synthesis. 
Multi-agent architectures~\citep{sami2024system, rouzrokh2025lattereview} have been wildly used, and decompose the pipeline into specialized agent roles, mimicking human editorial workflows.
Representatively, AutoSurvey~\citep{wang2024autosurvey} introduces a retrieval-outline-generation sequence that produces entire surveys via section-wise prompting. SurveyForge~\citep{yan2025surveyforge} extends this with memory modules and outline heuristics, aiming to enforce consistency across segments. SurveyX~\citep{liang2025surveyx} scales this further by relying on larger models and more complex pipeline, producing more robust and strict step-by-step outputs. 

Despite these advances, many systems adopt a static and compartmentalized approach. Outlines are typically fixed in advance, with no capacity to revise structure based on intermediate content. 
Subsections are often generated in parallel, lacking shared context, which weakens narrative flow and increases repetition or terminology drift. Citation usage also remains surface-level: references are selected from top-ranked snippets without tracing citation chains or enabling expansion~\citep{kasanishi2023scireviewgen}. 
In response, our work views survey writing as a dynamic process, one that requires adaptive planning, context-aware memory, and citation-traced retrieval. By continuously refining structural plans, maintaining cross-section consistency, and grounding generation in citation chains, we move toward more coherent and adaptive scientific surveys.

\section{Methodology}

\begin{figure*}[!t]
    \centering
    \includegraphics[width=0.98\textwidth]{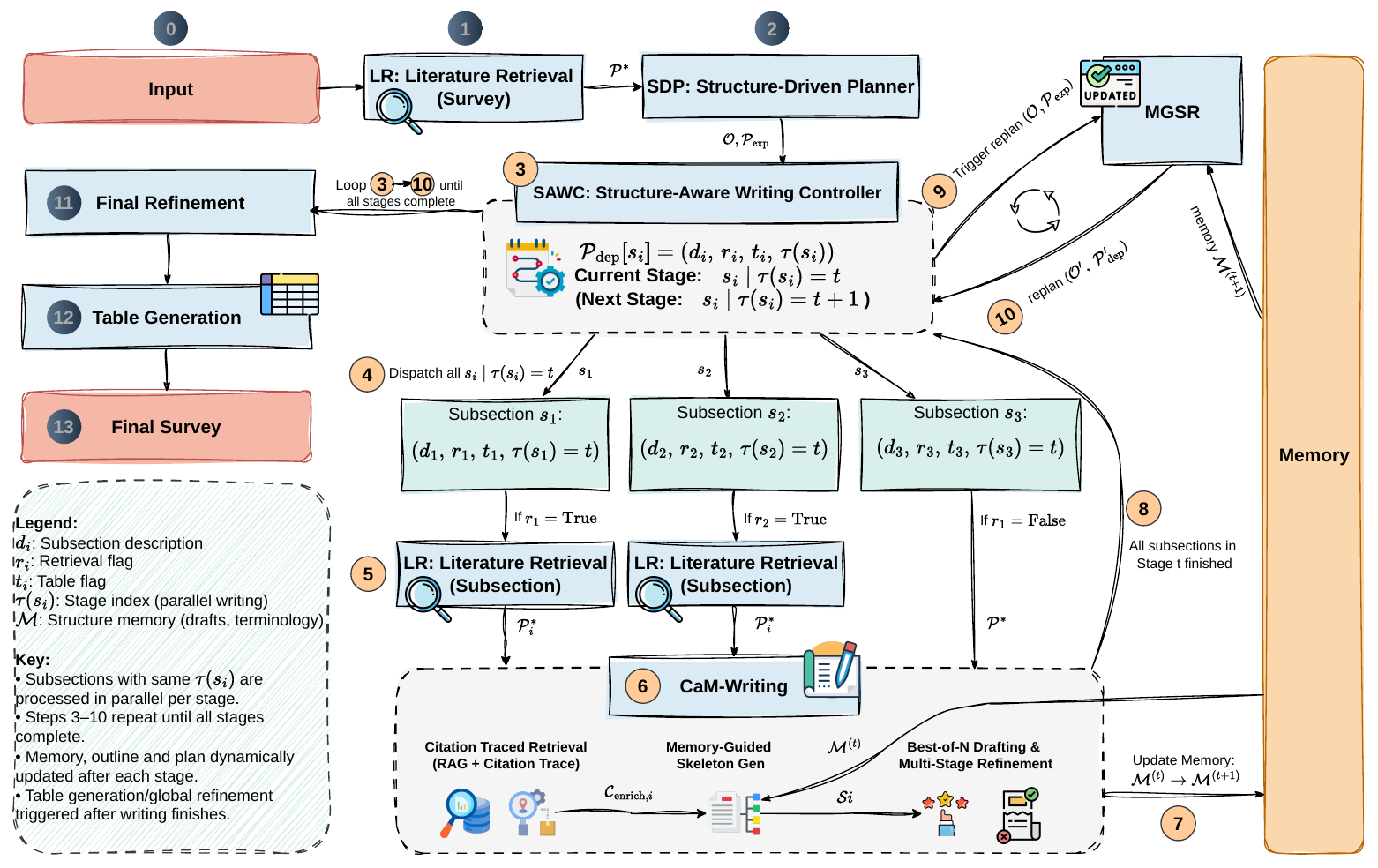}

\caption{
Overview of the \textbf{SurveyGen-I} pipeline for automatic academic survey generation. 
The system comprises three stages: (1) multi-stage \textbf{Literature Retrieval} (LR); 
(2) \textbf{PlanEvo}, a structure planning module integrating SDP (planning), SAWC (scheduling), and MGSR (dynamic replanning); 
(3) \textbf{CaM-Writing} for citation-aware subsection generation. 
Final refinement and table generation are performed after writing. Memory $\mathcal{M}$ accumulates writing content and terminology across stages to guide planning and ensure consistency.
}

    \label{fig:pipeline}
\end{figure*}

In this section, we propose \textbf{SurveyGen-I}, a novel framework for automatic survey generation. As shown in Figure~\ref{fig:pipeline}, it consists of three key stages: (1) \textbf{Literature Retrieval (LR)} performs multi-stage literature retrieval at both survey and subsection levels. (2) \textbf{Structure Planning with Dynamic Outline Evolution (PlanEvo)} generates a hierarchical outline and a dependency-aware writing plan, and dynamically updates both during generation to ensure cross-subsection consistency of the survey. (3) \textbf{CaM-Writing} generates each subsection with strong content consistency and rich citation coverage, combining citation-traced retrieval to recover influential references, memory-guided skeleton planning for content consistency, and best-of-$N$ draft selection to ensure high-quality generation.

\subsection{LR: Literature Retrieval}

\begin{figure*}[t]
    \centering
    \includegraphics[width=1\textwidth]{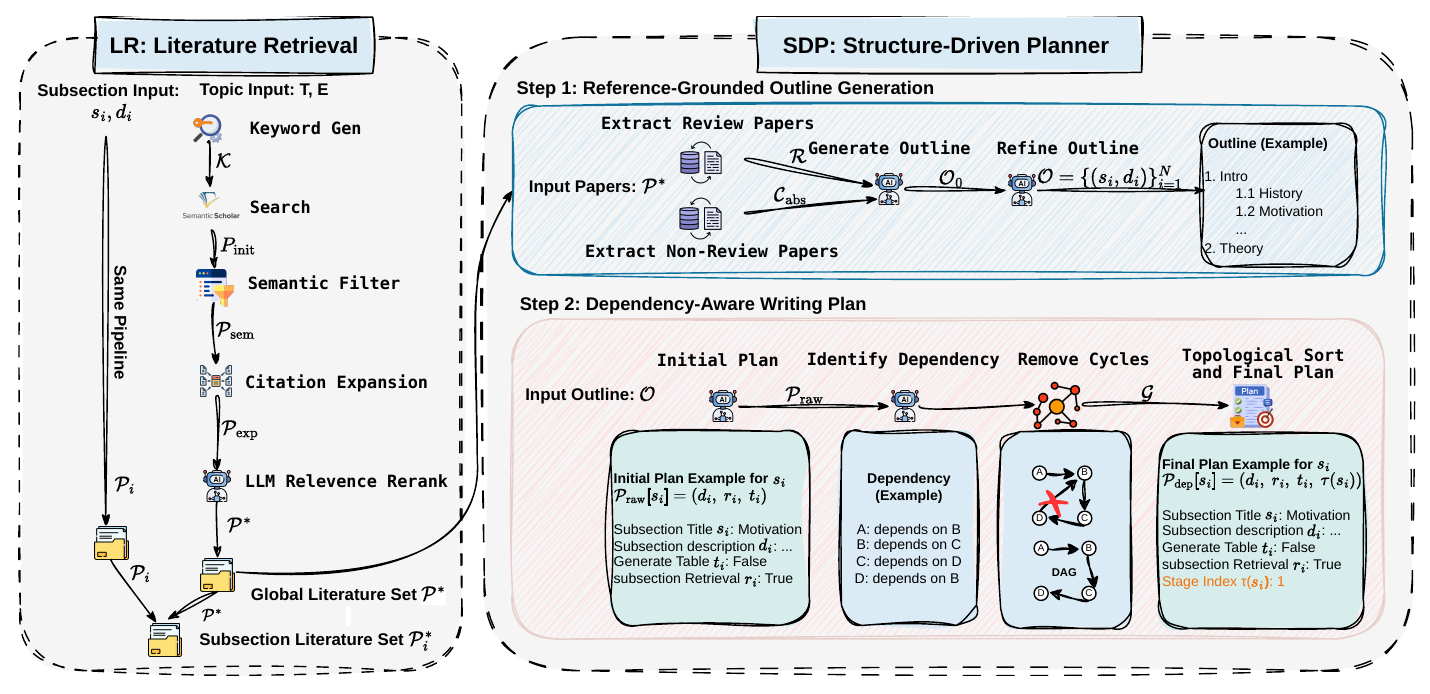}
    \caption{%
Details of the \textbf{Literature Retrieval} and \textbf{Structure-Driven Planner} components in \textbf{SurveyGen-I}. }

    \label{fig:lr_sdp}
\end{figure*}
To ensure that the generated survey is grounded with the most relevant and comprehensive research, our system adopts a multi-stage literature retrieval strategy that operates at both the survey and subsection levels. As shown in Figure~\ref{fig:pipeline}, this retrieval process provides the reference foundation for both structure planning (SDP; see Sec.~\ref{SDP}) and writing (CaM-Writing; see Sec.~\ref{sec:writing_agent}). The overall workflow is shown in Figure~\ref{fig:lr_sdp}; implementation details are provided in Appendix~\ref{retrieval}.

\subsubsection{Survey-Level Retrieval for Structure Planning}
\label{globalre}
For survey-level literature retrieval, an LLM is first prompted to generate a keyword set $\mathcal{K}$ based on the input topic $T$ and its description $E$ (see prompt in Figure~\ref{fig:key_word_generation_prompt}, Appendix~\ref{appendix:prompts}). These keywords are used to query Semantic Scholar \citep{ammar2018construction}, producing an initial set of papers $\mathcal{P}_{\text{init}}$. While keyword-based retrieval offers broad initial coverage, it may include irrelevant papers. To enhance topical precision, a semantic filtering step is applied. Specifically, both the input $(T, E)$ and each paper abstract $a_i$ are embedded using the \texttt{all-mpnet-base-v2} sentence transformer \citep{song2020mpnet}. Candidate papers with high cosine similarity to the input $(T, E)$ are retained, yielding a refined set $\mathcal{P}_{\text{sem}}$:
\begin{equation}
\mathcal{P}_{\text{sem}} = \{p_i \in \mathcal{P}_{\text{init}} \mid \cos(\mathbf{e}_{T,E}, \mathbf{e}_{a_i}) \geq \theta \}.
\end{equation}

To improve coverage and avoid missing influential work, $\mathcal{P}_{\text{sem}}$ is expanded by retrieving references and citations of top-ranked papers. These expanded papers are again filtered by embedding similarity for topical relevance. Finally, an LLM-based relevance scorer assesses all remaining papers with respect to $(T, E)$, and the top-ranked literature $\mathcal{P}^*$ is retained to support outline generation (see prompt template in Figure~\ref{fig:relevance_scoring_prompt}, Appendix~\ref{appendix:prompts}).

\subsubsection{Subsection-Level Retrieval for Writing}
\label{subsectionre}
In addition to survey-level retrieval for structure planning, subsection-level retrieval is optionally triggered during writing. For each subsection $s_i$ with its description $d_i$, a focused paper set $\mathcal{P}_i$ is constructed using the same retrieval pipeline as above, with $(s_i, d_i)$ as input. Whether this step is performed is controlled by a retrieval flag $r_i$ in the dependency-aware writing plan (see Sec.~\ref{SDP}). The final paper set used for writing each subsection is the combination of the survey-level set $\mathcal{P}^*$ and the subsection-level set $\mathcal{P}_i$, forming the combined paper set $\mathcal{P}^*_i$. This paper set captures both the global scope of the survey and the specific focus of each subsection.

\subsection{PlanEvo: Structure Planning with Dynamic Outline Evolution}

In this section, we introduce \textbf{PlanEvo}, a planning-centric framework for scalable and coherent survey outline generation and writing plan construction. PlanEvo consists of three tightly integrated components: the Structure-Driven Planner (SDP), the Structure-Aware Writing Controller (SAWC), and the Memory-Guided Structure Replanner (MGSR). Detailed designs for each component are presented in Sec.~\ref{SDP}, Sec.~\ref{SAWC}, and Sec.~\ref{MGSR}.

\subsubsection{SDP: Structure-Driven Planner}
\label{SDP}

The SDP module serves as the entry point of PlanEvo, transforming a specific research topic $(T, E)$ into a structured, executable plan that guides the full survey generation process. The overall workflow is shown in Figure~\ref{fig:lr_sdp}.

\paragraph{Reference-Grounded Outline Generation.}

A literature-grounded outline is essential for generating a coherent and well-structured survey. To build such an outline, the system first identifies review articles within the survey-level literature set $\mathcal{P}^*$ by analyzing metadata such as publication type. The structural outlines of these reviews $\mathcal{R}$ are then extracted from their full texts using LLMs and used as representative structural patterns to inspire the design of new outlines. The system then collects titles and abstracts of non-review papers in $\mathcal{P}^*$ to form the abstract-level content set $\mathcal{C}_\text{abs}$, which is then combined with $\mathcal{R}$ into a composite context $\mathcal{C}$. Given $\mathcal{C}$ and $(T, E)$, an LLM is prompted to generate an initial outline $\mathcal{O}_0$:
\begin{equation}
\mathcal{O}_0 = \{(s_i, d_i)\}_{i=1}^{N}
\end{equation}

where each subsection heading $s_i$ is paired with a brief description $d_i$ to provide more detailed guidance for writing subsections. To improve coherence and reduce redundancy, the initial outline $\mathcal{O}_0$ is refined by an LLM, yielding the final outline $\mathcal{O}$.

\paragraph{Dependency-Aware Writing Plan.}
To enable logically coherent and coordinated writing across subsections, we construct a dependency-aware writing plan $\mathcal{P}_{\text{dep}}$ based on the survey outline $\mathcal{O}$.

First, an initial plan $\mathcal{P}_{\text{raw}}$ is generated by prompting an LLM with $\mathcal{O}$. For each subsection $s_i$ with its description $d_i$, the plan specifies two control signals: whether additional literature retrieval is required ($r_i$), and whether a comparative table should be generated ($t_i$). These signals guide downstream tasks in subsection-level literature retrieval and table generation.
Each subsection takes the form:
\begin{equation}
\mathcal{P}_{\text{raw}}[s_i] = (d_i,\; r_i,\; t_i),
\end{equation}

Next, a structural dependency graph $\mathcal{G}_{\text{raw}} = (\mathcal{V}, \mathcal{E})$ is constructed by prompting an LLM to identify, for each subsection, its prerequisite subsections (see prompt in Figure~\ref{fig:writing_dependency_graph_construction_prompt}). Here, $\mathcal{V}$ includes all subsections $s_1, \dots, s_n$, and $\mathcal{E}$ contains edges $(s_i \rightarrow s_j)$ if $s_i$ is judged to be a prerequisite of $s_j$. Cycles are resolved by removing one edge per cycle, yielding a Directed Acyclic Graph (DAG) $\mathcal{G}$.

A topological sort is then applied to $\mathcal{G}$ to determine the writing order. Each subsection $s$ is assigned a \emph{stage index} $\tau(s)$, representing the length of the longest path ending at $s$ in the DAG:
\begin{equation}
\tau(s) =
\begin{cases}
0 & \text{if } \mathrm{In}(s) = \emptyset, \\
\max\limits_{s' \in \mathrm{In}(s)} \tau(s') + 1 & \text{otherwise}.
\end{cases}
\end{equation}

The final plan $\mathcal{P}_{\text{dep}}$ extends $\mathcal{P}_{\text{raw}}$ by attaching stage indices to each subsection:
\begin{equation}
\mathcal{P}_{\text{dep}}[s_i] = (d_i,\; r_i,\; t_i,\; \tau(s_i)).
\end{equation}
Subsections assigned the same stage index can be written in parallel, enabling multi-stage scheduling.

\subsubsection{SAWC: Structure-Aware Writing Controller}
\label{SAWC}
The SAWC module serves as the central orchestration engine across the entire writing process in the SurveyGen-I pipeline. 
Rather than being a single step, SAWC coordinates a sequence of interdependent modules, including writing stage scheduling (Step 4), subsection-level literature retrieval (Step 5), citation-aware writing (Step 6), memory updating (Step 7), dynamic structure replanning (Steps 9–10), global consistency refinement (Step 11), and table generation (Step 12). Its control flow is illustrated throughout the center path of Figure~\ref{fig:pipeline}.

\paragraph{Parallel Subsection Execution.} 
SAWC executes the dependency-aware writing plan $\mathcal{P}{\text{dep}}$ by activating all subsections with the same writing stage index $\tau(s_i)$ in parallel (Step 4). For each active subsection $s_i$, SAWC first checks the retrieval control flag $r_i$ in $\mathcal{P}_{\text{dep}}[s_i]$. If retrieval is required, SAWC triggers subsection-level literature retrieval (Step 5; see also Sec.~\ref{subsectionre}). The resulting paper set $\mathcal{P}_i^*$ is passed to the writing module (Step 6; see Sec.~\ref{sec:writing_agent}) for citation-aware subsection generation.

\paragraph{Memory Mechanism for Global Consistency.}

To ensure structural coherence and terminological consistency across the survey, SAWC maintains a dynamic \textbf{structure memory} $\mathcal{M}$ throughout writing. After each subsection $s_i$ is written (Step 6), the system extracts key domain-specific terminology using LLMs, and stores both the terminology and the draft content into $\mathcal{M}$ (Step 7). This accumulated memory is then used to (1) guide subsequent subsection writing by enforcing consistency (see Sec.~\ref{Skeleton-guided Writing with Memory Integration}), and (2) provide feedback for dynamic updates of the outline $\mathcal{O}$ and writing plan $\mathcal{P}_{\text{dep}}$ during structure replanning (see Sec.~\ref{MGSR}).

\paragraph{Dynamic Structure Refinement.}
At the end of each writing stage, which corresponds to the completion of all subsections $s_i$ with the same stage index $\tau(s_i)$, SAWC triggers the Memory-Guided Structure Replanner (MGSR; see Sec.~\ref{MGSR}) to revise the outline and the writing plan based on the accumulated memory $\mathcal{M}$ (Step 9–10). This stage-wise feedback loop ensures that structural adjustments are continuously informed by prior writing outputs before the next stage begins.

\paragraph{Final Refinement and Table Generation.} 
After all subsections are written, SAWC performs a final refinement step to improve global coherence (Step 11). An LLM analyzes the full draft to detect logical contradictions, redundancy, and terminological/style inconsistencies. Based on this diagnosis, the system rewrites affected subsections to ensure consistency. Then, for each subsection with the table flag $t_i$ enabled in $\mathcal{P}_{\text{dep}}[s_i]$, SAWC generates a structured table based on the retrieved paper set (Step 12). See Appendix \ref{appendix:tables} for details.

\subsubsection{MGSR: Memory-Guided Structure Replanner}
\label{MGSR}

After each writing stage, the MGSR module performs dynamic refinement of the outline and writing plan based on the accumulated memory $\mathcal{M}$ and the current outline $\mathcal{O}$. MGSR prompt an LLM (see prompt in Figure~\ref{fig:structural_revision_planning_for_unwritten_subsections_prompt}) to analyze redundancy, missing conceptual gaps, or suboptimal ordering within the unwritten subsections. It produces a set of structured revision actions (merge, delete, rename, reorder, add) applied to the remaining outline. The updated writing plan $\mathcal{P}_{\text{dep}}'$ is then derived from the revised outline $\mathcal{O}'$ through the same dependency-aware planning method used in the initial plan $\mathcal{P}_{\text{dep}}$ (see Sec.~\ref{SDP}). This enables memory-guided structural evolution throughout writing, ensuring that later sections are adaptively optimized based on prior content while maintaining global consistency.

\subsection{CaM-Writing: Citation-Aware Subsection Writing with Memory Guidance}
\label{sec:writing_agent}

This section introduces \textbf{CaM-Writing}, a citation-aware, memory-guided writing pipeline for generating each survey subsection. The pipeline integrates citation-traced retrieval to enhance literature coverage and citation diversity, skeleton-based generation guided by accumulated memory $\mathcal{M}$ to ensure content consistency, and multi-stage refinement to improve clarity, coherence, and citation integrity.

\subsubsection{Context Construction with Citation Tracing}

To construct a rich and contextually relevant evidence set for writing each subsection $s_i$ with description $d_i$, a RAG step is first applied over the contextual paper set $\mathcal{P}^*_i$, which includes both survey-level and subsection-specific literature. Top-ranked passages are selected to form the initial writing context $\mathcal{C}_{\text{rag}, i}$. However, the retrieved passages from academic papers often contain indirect citations such as "[23]" and "Ge et al., 2023". These citations typically refer to influential prior work that is not directly included in the retrieved documents. If the system relies solely on these secondary mentions without further resolution, it may overlook foundational or highly relevant papers. 

To address this, we introduce a \textbf{citation-tracing mechanism} that identifies such citations in $\mathcal{C}_{\text{rag}, i}$ and uses an LLM to determine whether each refers to an original source of a key concept or result. Traceable citations are resolved via the Semantic Scholar API, and their abstracts are appended to the base context $\mathcal{C}_{\text{rag}, i}$, forming the citation-enriched context $\mathcal{C}_{\text{enrich}, i}$. To maintain traceability, each enriched abstract is linked back to the original passage that cited it, allowing the writing model to understand the relationship between the mention and its source. For example, consider the passage:
\begin{quote}
“... recent work has introduced reward-balanced fine-tuning for alignment (Ge et al., 2023), showing improvements over DPO and RLHF ...”
\end{quote}
The LLM flags “(Ge et al., 2023)” as traceworthy, identifying it as introducing a core method. The system then resolves it to:
\begin{quote}
\textbf{Title:} Preserve Your Own Correlation: A New Reward-Balanced Fine-Tuning Method \\
\textbf{Abstract:} `We introduce a reward-balanced fine-tuning (RBF) framework for language model alignment...''
\end{quote}
This abstract is appended to the context, enabling the system to cite this traced paper directly in the generation. Further implementation details are provided in Appendix~\ref{appendix:citation_tracing}.

\subsubsection{Memory-Aligned Skeleton-Guided Generation}

\label{Skeleton-guided Writing with Memory Integration}

Given the enriched context $\mathcal{C}_{\text{enrich}, i}$, subsection title and description $(t_i, d_i)$, and the accumulated structure memory $\mathcal{M}$, the system first uses an LLM (see prompt in Figure~\ref{fig:wirting_skeleton_generation_conditioned_on_structure_memory_prompt}) to generate a writing skeleton $\mathcal{S}_i$ outlining the key conceptual points. The memory $\mathcal{M}$, which includes prior subsections and extracted terminology, ensures content coherence and terminology consistency across the survey.

\paragraph{Best-of-N Selection.} 
$N$ candidate drafts are first generated based on the subsection title $s_i$, description $d_i$, writing skeleton $\mathcal{S}_i$, and enriched context $\mathcal{C}_{\text{enrich}, i}$ (see prompt in Figure~\ref{fig:full_subsection_writing_with_rag_and_citation_tracing_prompt}). An LLM then evaluates the candidates and selects the best version based on alignment with the skeleton, contextual relevance, and overall writing quality.

\paragraph{Subsection-Level Refinement.}
To further improve the selected draft, a three-stage refinement is applied.
First, the structure is adjusted to better reflect the conceptual flow defined by $\mathcal{S}i$.
Second, the draft undergoes citation refinement, where the LLM rewrites the text based on $\mathcal{C}_{\text{enrich}, i}$.
Finally, the draft is polished to enhance fluency and clarity.

\label{sec:final_refine}

\section{Experiments and Results}

\subsection{Evaluation Setup.}

We compare SurveyGen-I with three representative baselines: 
AutoSurvey~\citep{wang2024autosurvey}, SurveyForge~\citep{yan2025surveyforge}, and SurveyX~\citep{liang2025surveyx}. 
We collect demo reports from SurveyForge and SurveyX official project pages. 
For AutoSurvey, we generate reports on matched topics for a fair comparison and used the same model used in SurveyGen-I, with GPT4o-mini~\citep{openai2024gpt4omini}. 
We also construct a new benchmark covering six major scientific domains, each with $\sim$30 subtopics.

\subsection{Evaluation Metrics}

We comprehensively evaluate \textit{SurveyGen-I} against three competitive baselines across two core dimensions, content quality and reference quality.

\paragraph{Content Quality Evaluation.}

Measures the structural and semantic strength of the generated survey. This includes five sub-dimensions: \textit{coverage, relevance, structure, synthesis, and consistency}. Each aspect is scored by LLM-as-Judge~\citep{li2024llms} (specifically, rated by GPT4o-mini), with explanation-based prompts to reduce variance. This directly reflects the impact of our MGSR and CaM-Writing, which aim to improve global coherence, abstraction, and flow.
Evaluation criteria can be found in the appendix~\ref{appendix:prompts}. We compute the final content quality score (CQS) as the average of five evaluation dimensions.

\paragraph{Reference Quality Evaluation.} 
To assess the effectiveness and recency of reference usage in the generated survey, we adopt three reference-level metrics that reflect citation coverage, intensity, and timeliness. 
The Number of References (NR) counts the distinct cited works, measuring the breadth of literature coverage. 
The Citation Density (CD) computes the number of unique citation markers per character of text (excluding the reference section), reflecting how frequently references are integrated into the main narrative. For reporting clarity, we scale CD by a factor of $10^4$. 
The Recency Ratio (RR@k) measures the proportion of all cited references that were published within a recent time window (e.g., within the past k=3 years). A higher RR indicates better engagement with the latest developments in the field, and reflects the model's ability to retrieve and integrate timely literature.

\subsection{Main Results}

We report evaluation results across content quality and reference behavior, along with an ablation-based component analysis. 
\textit{SurveyGen-I} is compared against three state-of-the-art baselines: AutoSurvey, SurveyX, and SurveyForge. Our results demonstrate that \textit{SurveyGen-I} achieves significant improvements across all dimensions, showing its effectiveness for automated survey generation.

\paragraph{Content Quality.}
SurveyGen-I achieves consistent improvements across all five content quality dimensions compared to prior systems (Table~\ref{tab:report_metrics}). The overall score reaches 4.59, outperforming the best baseline (SurveyForge: 4.23) by +0.36. Largest gains are observed in structural flow (STRUC: +0.21) and synthesis (SYN: +0.41), indicating that the model maintains a coherent narrative while integrating information from diverse sources. Coverage (4.72) and relevance (4.76) also lead all baselines, suggesting high topical breadth and alignment. Consistency (4.59) improves notably over SurveyX (4.29), reflecting stability in terminology and phrasing across sections. 
Notably, SurveyX uses GPT4o~\cite{hurst2024gpt}, whereas our system relies on a smaller and more cost-efficient model, making the performance gap especially significant. 
Observed quality gains suggest that systems combining structural adaptivity, iterative refinement, and citation-tracing can more reliably generate coherent and well-grounded surveys.

\begin{table}[htbp]
\centering
\resizebox{1\linewidth}{!}{
\begin{tabular}{lrrrrrr}
    \hline
    \textbf{Model} & \textbf{Overall~$\uparrow$} & \textbf{Cov~$\uparrow$} & \textbf{Rel~$\uparrow$} & \textbf{Struc~$\uparrow$} & \textbf{Syn~$\uparrow$} & \textbf{Consis~$\uparrow$} \\
    \hline
    AutoSurvey & 4.08 & 4.10 & 4.17 & 4.03 & 4.10 & 4.00 \\
    SurveyX & 4.13 & 4.10 & 4.33 & 4.00 & 3.95 & 4.29 \\
    SurveyForge & 4.23 & 4.31 & 4.41 & 4.07 & 4.21 & 4.17 \\
    \rowcolor{SkyBlue!10}\textbf{Ours} & \textbf{4.59} & \textbf{4.72} & \textbf{4.76} & \textbf{4.28} & \textbf{4.62} & \textbf{4.59} \\
    \hline
  \end{tabular}
  }
  \caption{\label{tab:report_metrics} LLM-based evaluation scores across multiple survey quality dimensions. 
  Higher scores reflect better coverage (COV), relevance (REL), structural flow (STRUC), synthesis (SYN), and consistency (CONSIS). SurveyGen-I leads across all.}
\end{table}

\begin{figure}[t]
  \centering
  \begin{subfigure}{0.48\linewidth}
    \includegraphics[width=\linewidth]{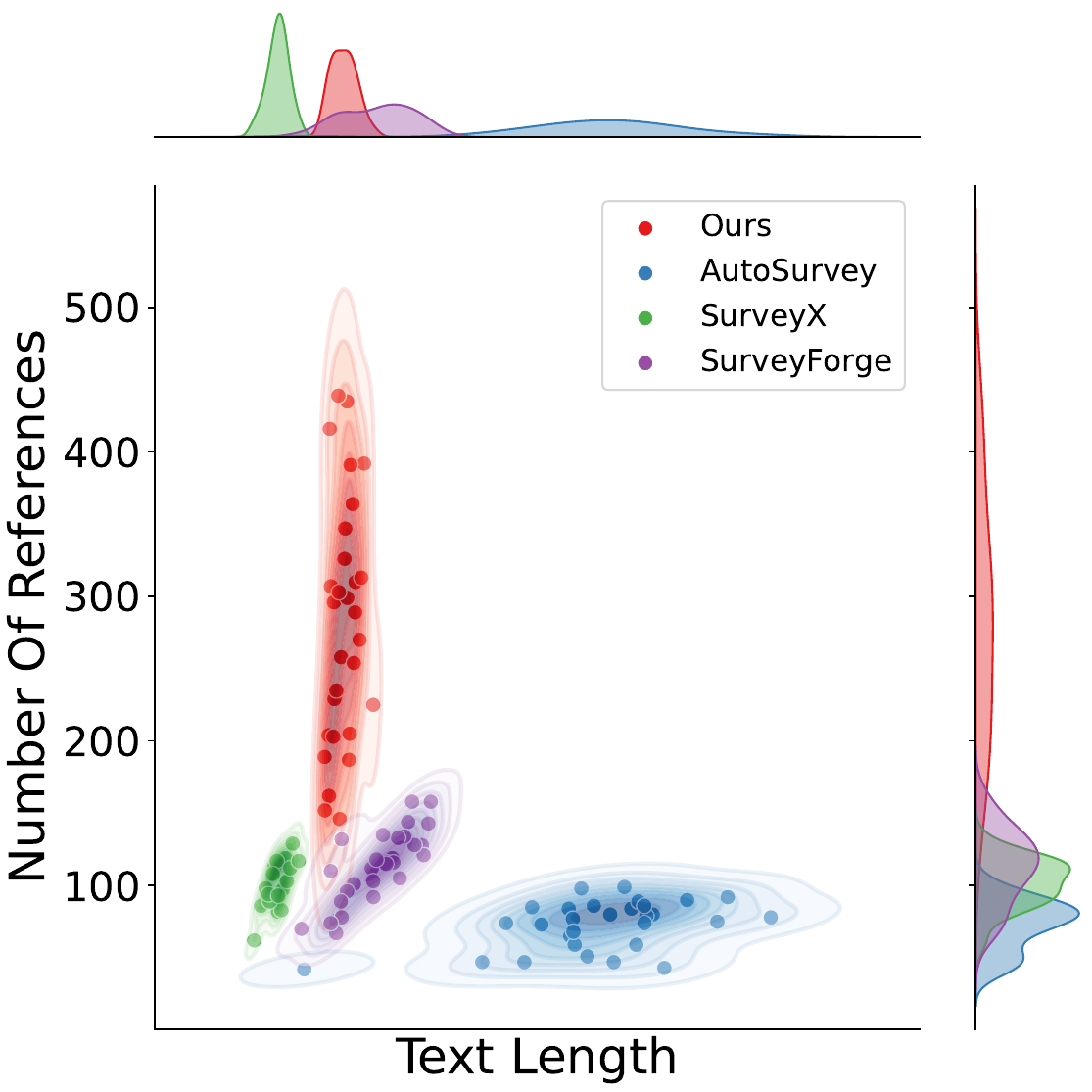}
    \caption{Ref count vs. length}
    \label{fig:citation_behavior_a}
  \end{subfigure}
  \hfill
  \begin{subfigure}{0.48\linewidth}
    \includegraphics[width=\linewidth]{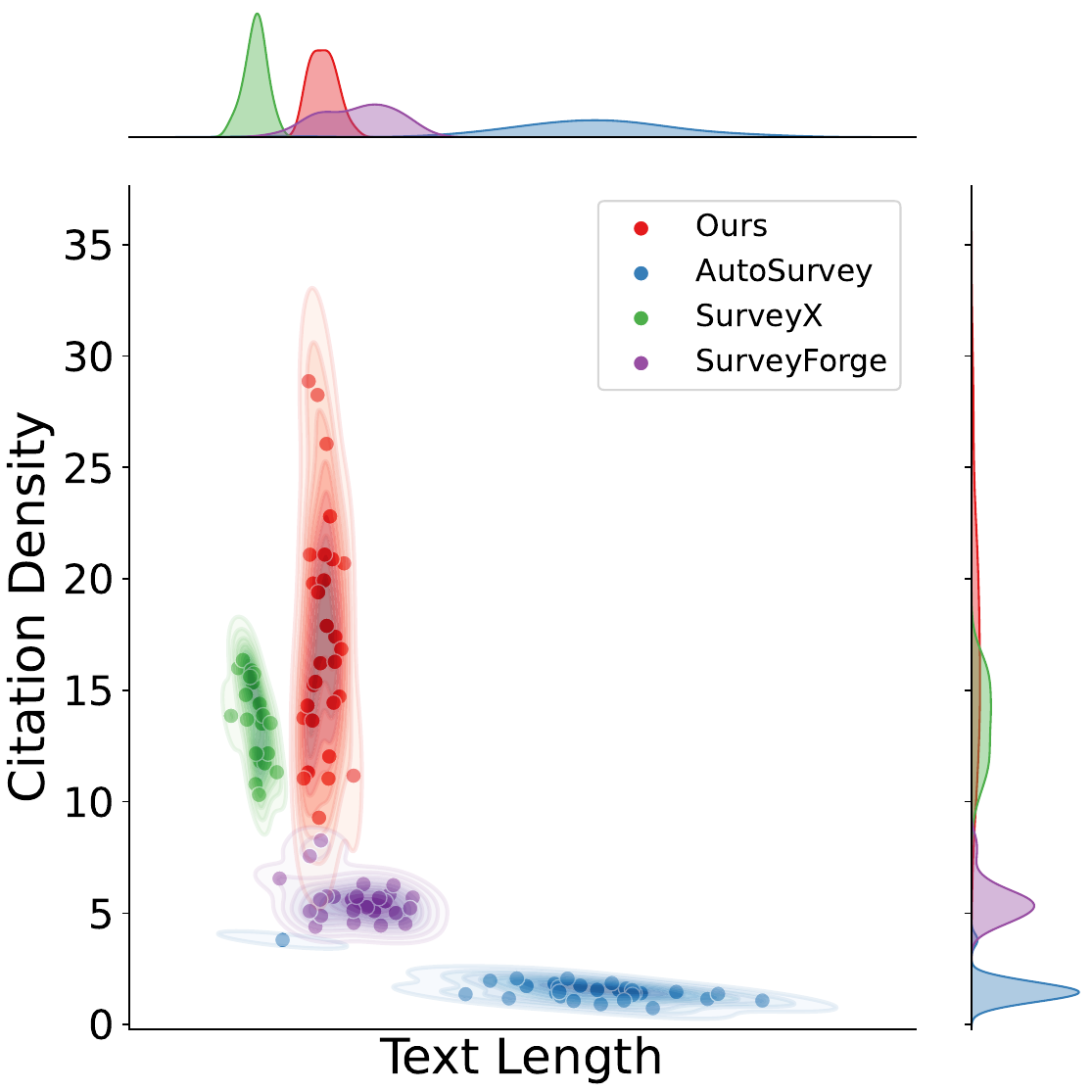}
    \caption{Citation density}
    \label{fig:citation_behavior_b}
  \end{subfigure}
  \vspace{-1ex}
  \caption{
    Citation behavior comparisons across models using KDE-enhanced scatter plots. 
    \textbf{(a)} Number of references vs. text length. 
    SurveyGen-I demonstrates a steeper citation scaling curve, suggesting deeper integration of references even in longer texts.
    \textbf{(b)} Citation density vs. text length. SurveyGen-I maintains denser citation patterns across all lengths.
  }
  \label{fig:citation_behavior_two_plots}
  \vspace{-1.5ex}
\end{figure}

\paragraph{Reference Quality.}
In terms of citation quality and scientific grounding, SurveyGen-I exhibits both broader and denser reference usage. It cites 281 unique works per survey on average (Table~\ref{tab:combined_recency_citation}), representing a sharp increase over SurveyX (102) and AutoSurvey (73). Citation density also rises substantially (17.28), exceeding SurveyForge (5.52) by around 3 times, indicating tighter integration of references into the body text.
Importantly, 89.1\% of all citations are published within the past 5 years (RR@5), compared to 66.7\% in SurveyX and SurveyForge, demonstrating significantly improved recency alignment. 

The steep scaling trend between reference count and text length in Figure~\ref{fig:citation_behavior_a} shows that text length remains relatively stable in SurveyGen-I, reflecting the fixed-length constraint imposed during generation. It also demonstrates that SurveyGen-I includes the most references overall. 
In contrast, AutoSurvey consistently generates fewer references, and its citation count remains relatively flat, even as its text length slightly increases, which is unexpected given that the length parameter was controlled across all generations. This suggests weaker responsiveness to contextual expansion and underutilization of available content space. 
Figure~\ref{fig:citation_behavior_b} further shows that SurveyGen-I consistently maintains a high citation density across varying text lengths, indicating robust integration of information-dense content.
\begin{table}[htbp]
  \centering
  \resizebox{1\linewidth}{!}{
  \begin{tabular}{lrrrrrrr}
    \hline
    \textbf{Model} & \textbf{RR@1} & \textbf{RR@3} & \textbf{RR@5} & \textbf{RR@7} & \textbf{RR@10} & \textbf{CD}  & \textbf{NR} \\
    \hline
    AutoSurvey & 0.174 & 0.639 & 0.837 & 0.940 & \textbf{0.992} & 1.54 & 73 \\ 
    SurveyX & 0.239 & 0.484 & 0.667 & 0.792 & 0.916 & 13.57 & 102 \\
    SurveyForge & 0.137 & 0.437 & 0.667 & 0.824 & 0.907 & 5.52 & 113 \\
    \rowcolor{SkyBlue!10}\textbf{Ours} & \textbf{0.478} & \textbf{0.759} & \textbf{0.891} & \textbf{0.955} & \textbf{0.985} & \textbf{17.28} &  \textbf{281} \\
    \hline
  \end{tabular}
  }
  \caption{\label{tab:combined_recency_citation} Performance comparison across models based on recency-focused citation behavior and structural citation metrics. RR@k indicates the proportion of recent references among the top-k citations. CD (scaled by \( \times 10^4 \)) measures citation density, and NR denotes the total number of cited references.}
\end{table}

\paragraph{Ablation Analysis.}
Table~\ref{tab:ablation} reports the impact of removing specific behaviors from SurveyGen-I. The full model yields the highest overall score (4.77), with synthesis and structure both at 4.86 and 4.71, respectively. Disabling final refinement results in the steepest quality drop (Overall: --0.43), particularly in synthesis (--0.43) and structure (--0.57), indicating that single-pass generation without revision is insufficient for maintaining narrative integration. Fixed planning further reduces structural flow (STRUC: --0.42) and consistency (CONSIS: --0.28), suggesting that static outlines limit the model’s ability to adjust to unfolding content. Removing citation resolution reduces the number of distinct references by 61 and lowers relevance by 0.29, despite stable consistency.

\begin{table}[htbp]
\centering
\resizebox{1\linewidth}{!}{
\begin{tabular}{lrrrrrrr}
    \hline
    \textbf{Model} & \textbf{Overall~$\uparrow$} & \textbf{Cov~$\uparrow$} & \textbf{Rel~$\uparrow$} & \textbf{Struc~$\uparrow$} & \textbf{Syn~$\uparrow$} & \textbf{Consis~$\uparrow$} & \textbf{NR~$\uparrow$}\\
    \hline

    Ours (w/o Citation Trace) & 4.60 & 4.57 & 4.57 & 4.43 & 4.71 & \textbf{4.71}  & 225\\
    Ours (w/o Plan Update) & 4.49 & 4.57 & 4.71 & 4.29 & 4.43 & 4.43 & 212\\
    Ours (w/o Refine) & 4.34 & 4.43 & 4.43 & 4.14 & 4.43 & 4.29 & \textbf{286} \\
        \rowcolor{SkyBlue!10}\textbf{Ours (Full)} & \textbf{4.77} & \textbf{4.71} & \textbf{4.86} & \textbf{4.71} & \textbf{4.86} & \textbf{4.71} & \textbf{286} \\
    \hline
    
  \end{tabular}
  }
\caption{\label{tab:ablation} Evaluation results of ablation variants. Each component (Trace, Plan Update, Refine) contributes to overall quality.}
\end{table}
\section{Conclusion}

We present SurveyGen-I, a fully automated framework for generating academic surveys with high consistency, citation coverage, and structural coherence. By integrating multi-level retrieval, adaptive planning, and memory-guided writing, SurveyGen-I effectively captures complex literature landscapes and produces high-quality surveys without manual intervention. Extensive evaluations across six scientific domains demonstrate its effectiveness over existing methods, marking a step forward in reliable and scalable scientific synthesis.

\clearpage
\section*{Limitations}

While \textit{SurveyGen-I} shows consistently strong performance across benchmarks, our framework adopts an online retrieval strategy to ensure access to up-to-date literature. 
However, this design introduces network sensitivity, variable latency, and reliance on third-party APIs, which may restrict full-text access due to licensing constraints. 
Compared to offline-indexed corpora used in prior work, our approach trades retrieval speed and infrastructure control for broader coverage and freshness. 

Additionally, for niche or emerging topics with limited source material, the achievable survey length and depth are naturally constrained. 
This shows a general challenge in automatic survey generation: content quality is ultimately bounded by the availability and granularity of the source literature.
Moreover, some evaluation signals may reflect subjective preferences rather than universal writing standards. 
We anticipate that broader community adoption and feedback will help guide future enhancements.

\bibliography{references}
\clearpage
\appendix
\label{sec:appendix}

\section{SurveyGen-I Pipeline}
\begin{algorithm}[th]
\caption{SurveyGen-I Pipeline}
\label{alg:e2e-survey}
\small
\begin{algorithmic}[1]
\Require Research topic $T$, description $E$
\Statex \textbf{// LR: Multi-Stage Literature Retrieval}
\State $\mathcal{P}_{\text{init}} \gets \textsc{KeywordSearch}(T, E)$
\State $\mathcal{P}_{\text{sem}} \gets \textsc{SemanticFilter}(\mathcal{P}_{\text{init}}, T, E)$
\State $\mathcal{P}^* \gets \textsc{ExpandAndRank}(\mathcal{P}_{\text{sem}}, T, E)$

\Statex \textbf{// PlanEvo: Structure Planning and Scheduling}
\State $\mathcal{O} \gets \textsc{GenerateOutline}(\mathcal{P}^*, T, E)$
\State $\mathcal{P}_{\text{dep}} \gets \textsc{BuildSchedule}(\mathcal{O})$
\State $\mathcal{M} \gets \emptyset$ 

\Statex \textbf{// CaM-Writing}
\For{$t = 0$ to $\max\limits_{s_i \in \mathcal{O}} \tau(s_i)$}
    \State $\mathcal{S}_t \gets \{ s_i \in \mathcal{O} \mid \tau(s_i) = t \}$
    \ForAll{$s_i \in \mathcal{S}_t$}
        \State $(d_i, r_i, t_i) \gets \mathcal{P}_{\text{dep}}[s_i]$
        \If{$r_i = \texttt{True}$}
            \State $\mathcal{P}_i \gets \textsc{SubsectionRetrieval}(s_i, d_i)$
        \Else
            \State $\mathcal{P}_i \gets \emptyset$
        \EndIf
        \State $\mathcal{P}^*_i \gets \mathcal{P}^* \cup \mathcal{P}_i$ 
        \State $\mathcal{C}_{\text{rag}, i} \gets \textsc{RAGRetrieve}(s_i, \mathcal{P}^*_i)$
        \State $\mathcal{C}_{\text{enrich}, i} \gets \textsc{CitationTrace}(\mathcal{C}_{\text{rag}, i})$
        \State $\mathcal{S}_i \gets \textsc{MakeSkeleton}(s_i, d_i, \mathcal{M})$
        \State $\hat{s}_i^{(1:N)} \gets \textsc{Write}(s_i, d_i, \mathcal{S}_i, \mathcal{C}_{\text{enrich}, i})$
        \State $\hat{s}_i^* \gets \textsc{SelectBest}(\hat{s}_i^{(1:N)})$
        \State $\hat{s}_i \gets \textsc{Refine}(\hat{s}_i^*, \mathcal{S}_i, \mathcal{C}_{\text{enrich}, i})$
        \State $\mathcal{M} \gets \mathcal{M} \cup \textsc{ExtractMemory}(\hat{s}_i)$
    \EndFor
    \State $(\mathcal{O}', \mathcal{P}'_{\text{dep}}) \gets \textsc{\textbf{MGSR}.UpdatePlan}(\mathcal{O}, \mathcal{P}_{\text{dep}}, \mathcal{M})$
    \State $\mathcal{O} \gets \mathcal{O}'; \quad \mathcal{P}_{\text{dep}} \gets \mathcal{P}'_{\text{dep}}$
\EndFor

\State $\textsc{GlobalRefine}(\mathcal{O}, \mathcal{M})$
\State \Return Final survey draft
\end{algorithmic}
\label{alg}
\end{algorithm}

\section{SurveyGen-I Implementation Details}
\label{implementation}

\subsection{Implementation of Literature Retrieval}
\label{retrieval}

Our literature retrieval system differentiates between survey-level and subsection-level retrieval: survey-level retrieval is based on the user-specified topic and its explanation, which are clarified via LLM to produce a refined query capturing the overall survey scope, while subsection-level retrieval takes the section title, section description, subsection title, and subsection description as input, allowing the LLM to disambiguate local intent within the broader context. For filtering, we apply a two-stage thresholding strategy: papers are first filtered by cosine similarity (threshold 0.3) using \texttt{all-mpnet-base-v2} embeddings \cite{song2020mpnet}, then further filtered using an LLM-assigned relevance score, retaining only those with scores $\geq$ 70. These thresholds were chosen empirically to balance precision and recall across diverse topics. If no papers meet the relevance threshold, a fallback selects the top 5 most relevant candidates to ensure writing continuity. Additionally, we limit the maximum retained papers per query to 30 to control downstream processing cost. Citation-based expansion is only applied to top-scoring papers (typically the top 10).

All filtered papers are associated with structured metadata, including bibkey, title, abstract, paper ID, and download URL. We download each paper’s PDF asynchronously using available metadata from Semantic Scholar. The downloaded PDFs are parsed with PyMuPDF \cite{pymupdf2025} to extract clean textual content, which is stored alongside the metadata for downstream use in vectorstore construction, RAG retrieval, and citation tracing. All paper metadata and extracted content are persisted in CSV format, and BibTeX entries for retained papers are compiled into a unified reference file for LaTeX formatting.
\subsection{Outline and Writing Plan Generation}

\subsubsection{Survey Outline Example (Structured JSON)}
The following excerpt illustrates the JSON-based outline used to guide planning and writing. It defines high-level sections and their conceptual subsections:

\begin{lstlisting}[language=json]
{
  "section_title": "Fine-Tuning Methodologies for Enhanced Translation",
  "section_description": "This section categorizes and analyzes various fine-tuning methodologies employed in multilingual models, emphasizing their effectiveness for Chinese to Malay translation.",
  "subsections": [
    {
      "subsection_title": "Adaptive Fine-Tuning Techniques",
      "subsection_description": "Discusses adaptive fine-tuning methods, including Layer-Freezing and Low-Rank Adaptation, and their roles in optimizing model performance."
    },
    {
      "subsection_title": "Utilization of Adapter Modules",
      "subsection_description": "Explores the implementation of adapter modules for fine-tuning, highlighting their efficiency and effectiveness in multilingual speech translation tasks."
    },
    {
      "subsection_title": "Data Augmentation Strategies",
      "subsection_description": "Investigates the role of data augmentation techniques, including code-switching, in enhancing model robustness and translation quality."
    }
  ]
}
\end{lstlisting}

\subsubsection{Writing Plan Example (Execution-Ready Format)}
The following snippet is a real system-generated writing plan segment for the same section. It includes execution flags, dependency controls, and indexing metadata:

\begin{lstlisting}[language=json]
[
  {
    "section_title": "Fine-Tuning Methodologies for Enhanced Translation",
    "subsection_title": "Adaptive Fine-Tuning Techniques",
    "subsection_description": "Discusses adaptive fine-tuning methods, including Layer-Freezing and Low-Rank Adaptation, and their roles in optimizing model performance.",
    "index": 2,
    "trigger_additional_search": true,
    "generate_table": true,
    "depends_on": [
      "Challenges in Data Availability and Quality"
    ]
  },
  {
    "section_title": "Fine-Tuning Methodologies for Enhanced Translation",
    "subsection_title": "Utilization of Adapter Modules",
    "subsection_description": "Explores the implementation of adapter modules for fine-tuning, highlighting their efficiency and effectiveness in multilingual speech translation tasks.",
    "index": 3,
    "trigger_additional_search": true,
    "generate_table": true,
    "depends_on": [
      "Adaptive Fine-Tuning Techniques"
    ]
  },
  {
    "section_title": "Fine-Tuning Methodologies for Enhanced Translation",
    "subsection_title": "Data Augmentation Strategies",
    "subsection_description": "Investigates the role of data augmentation techniques, including code-switching, in enhancing model robustness and translation quality.",
    "index": 3,
    "trigger_additional_search": true,
    "generate_table": true,
    "depends_on": [
      "Challenges in Data Availability and Quality",
      "Adaptive Fine-Tuning Techniques"
    ]
  }
]
\end{lstlisting}

\subsubsection{Implementation Notes}

The \texttt{index} field determines batch-level parallel writing: subsections with the same index value can be written concurrently. The \texttt{depends\_on} field specifies logical dependencies between subsections, indicating which earlier subsections must be completed before the current subsection. The \texttt{trigger\_additional\_search} flag controls whether extra paper retrieval will be performed per subsection. Finally, \texttt{generate\_table} signals whether this subsection requires automatic generation of a literature table summarizing relevant methods, datasets, or evaluation metrics.

\subsection{Table Generation Implementation}
\label{appendix:tables}

\subsubsection{Overview}

For each subsection in the writing plan, the system determines whether to generate a table based on the number and diversity of relevant papers. If the number of filtered, non-review papers exceeds a certain threshold (typically 10), a \textit{method aggregation table} is created to group papers into conceptual categories. Otherwise, the system generates an \textit{aspect-based comparison table} that compares papers across a small set of important dimensions. 

\subsubsection{Method Aggregation Table}

When a subsection contains at least ten papers, the system generates a table that organizes the literature into high-level categories under a common comparative theme. The core aspect, for example, training paradigm, objective, or architecture type is first inferred using an LLM based on the subsection description and paper abstracts. A set of 4--6 concise method categories is then proposed by analyzing the previously written subsection content.

Each paper is assigned to one or more categories using RAG. The system constructs a query from the paper’s metadata, retrieves relevant content from the paper’s vectorized representation, and uses an LLM to classify the paper accordingly. Categories with fewer than two papers or labeled as ``Others'' are excluded from the final table.

\subsubsection{Aspect-Based Comparison Table}
If a subsection contains fewer than ten papers, the system generates a table comparing these papers across several dimensions, typically three to five, following the method proposed in \citet{newman2024arxivdigestables}. These aspects are automatically selected by analyzing the subsection's topic, abstract, and the written text of this subsection. For each (paper, aspect) pair, a dedicated query is constructed, and top-ranked passages are retrieved from the paper’s content using FAISS-based similarity search and reranking. These snippets are summarized using an LLM into short, informative values. The resulting table presents papers as rows and aspects as columns.

\subsection{Details of Citation Tracing for Context Grounding}
\label{appendix:citation_tracing}

\subsubsection{Subsection-Specific RAG Retrieval and Reranking}

Given a subsection title $s_i$ and its description $d_i$, a semantic query is issued to retrieve relevant literature passages. The retrieval corpus is the union of two sets: the survey-level paper collection $\mathcal{P}^*$ (covering the full survey scope) and the local set $\mathcal{P}_i$ specific to $s_i$. All papers are parsed and embedded using a BGE-based embedding model \cite{bge_embedding}, with metadata including \texttt{bibkey}, \texttt{title}, \texttt{abstract}, and \texttt{pdf\_path}.

For each source, top-$k$ documents are retrieved based on vector similarity. To ensure semantic precision, we apply a reranking step using a cross-encoder reranker (BGE-Reranker \cite{bge_embedding}). The reranker receives $(\text{query}, \text{document})$ pairs and assigns relevance scores, from which the top-10 documents are selected to form the initial RAG context $\mathcal{C}_{\text{rag}, i}$.

This hybrid retrieval strategy (dense + rerank) ensures high relevance, while combining survey-level and subsection-level sources ensures both broad coverage and local focus.

\subsubsection{Citation Marker Detection and Traceworthiness Evaluation}

Each passage in $\mathcal{C}_{\text{rag}, i}$ is scanned for citation markers using regular expressions. The system detects both numeric patterns such as “[17]” and author-year formats such as “(Ge et al., 2023)”. These markers are then evaluated for whether they refer to an original contribution worth tracing.

To perform this evaluation, we prompt an LLM with the subsection title, description, and raw passage text. The prompt asks the LLM to identify all citation markers and return a structured assessment for each, including a boolean traceworthiness flag and a natural language explanation. The output is parsed into a list of dictionaries; only citations flagged as traceworthy are considered for resolution.

\subsubsection{Citation Resolution and Traced Source Retrieval}

Once a citation marker is identified as traceworthy, the system attempts to resolve it to the original paper where the referenced idea or method was first introduced. This resolution yields a traced source, typically represented by its title, abstract, and bibkey. The abstract of the resolved paper is retrieved and appended to the RAG context, ensuring that subsequent generation relies on the original contribution rather than a potentially distorted secondary reference.

\section{AutoSurvey Implementation}
\label{appendix:autosurvey-config}

We follow an open-source AutoSurvey implementation with the generation model set to \texttt{gpt-4o-mini-2024-07-18}, the embedding model set to \texttt{nomic-ai/nomic-embed-text-v1} \cite{nussbaum2024nomic}, and retrieval configured with 2500 papers for outline construction. To ensure a fair comparison, the list of topics used for generation is aligned exactly with that of our system.

\section{Subtopic-Level Analysis of Recency Behavior}

\begin{table*}[ht]
  \centering
    \resizebox{0.7\linewidth}{!}{

  \begin{tabular}{l|l|r|r|r|r|r}
    \hline
    \textbf{Subtopic} & \textbf{Model} & \textbf{Rr@1} & \textbf{Rr@3} & \textbf{Rr@5} & \textbf{Rr@7} & \textbf{Rr@10} \\
    \hline
    AI Applications and Multidisciplinary Topics & AutoSurvey & 0.085 & 0.529 & 0.747 & 0.910 & \textbf{0.988} \\
    AI Applications and Multidisciplinary Topics & Ours & \textbf{0.586} & \textbf{0.811} & \textbf{0.901} & 0.938 & 0.971 \\
    AI Applications and Multidisciplinary Topics & SurveyForge & 0.125 & 0.361 & 0.638 & 0.802 & 0.900 \\
    AI Applications and Multidisciplinary Topics & SurveyX & 0.444 & 0.758 & 0.894 & \textbf{0.939} & 0.978 \\
    \hline
    Database and Big Data & AutoSurvey & 0.375 & \textbf{0.833} & \textbf{0.870} & \textbf{0.910} & \textbf{0.966} \\
    Database and Big Data & Ours & \textbf{0.544} & 0.749 & 0.862 & 0.893 & 0.939 \\
    Database and Big Data & SurveyForge & 0.104 & 0.279 & 0.445 & 0.512 & 0.613 \\
    Database and Big Data & SurveyX & 0.088 & 0.265 & 0.432 & 0.602 & 0.803 \\
    \hline
    Language Models & AutoSurvey & 0.349 & 0.793 & 0.925 & 0.967 & \textbf{0.997} \\
    Language Models & Ours & \textbf{0.519} & \textbf{0.855} & \textbf{0.936} & \textbf{0.977} & 0.993 \\
    Language Models & SurveyForge & 0.225 & 0.571 & 0.739 & 0.857 & 0.916 \\
    Language Models & SurveyX & 0.356 & 0.669 & 0.832 & 0.925 & 0.984 \\
    \hline
    Learning Algorithms and Foundations & AutoSurvey & 0.080 & 0.541 & 0.788 & \textbf{0.940} & \textbf{0.991} \\
    Learning Algorithms and Foundations & Ours & \textbf{0.328} & \textbf{0.614} & \textbf{0.806} & 0.922 & 0.977 \\
    Learning Algorithms and Foundations & SurveyForge & 0.060 & 0.322 & 0.579 & 0.803 & 0.900 \\
    Learning Algorithms and Foundations & SurveyX & 0.185 & 0.372 & 0.538 & 0.624 & 0.812 \\
    \hline
    Network, Systems, Infrastructure & AutoSurvey & 0.140 & 0.548 & 0.838 & 0.935 & \textbf{0.996} \\
    Network, Systems, Infrastructure & Ours & \textbf{0.504} & \textbf{0.717} & \textbf{0.872} & \textbf{0.966} & 0.994 \\
    Network, Systems, Infrastructure & SurveyForge & 0.109 & 0.364 & 0.619 & 0.781 & 0.932 \\
    Network, Systems, Infrastructure & SurveyX & 0.089 & 0.287 & 0.528 & 0.734 & 0.953 \\
    \hline
    Vision, Video, and Image Generation & AutoSurvey & 0.137 & 0.678 & 0.865 & 0.947 & 0.995 \\
    Vision, Video, and Image Generation & Ours & \textbf{0.472} & \textbf{0.803} & \textbf{0.937} & \textbf{0.982} & \textbf{0.998} \\
    Vision, Video, and Image Generation & SurveyForge & 0.145 & 0.534 & 0.763 & 0.901 & 0.945 \\
    Vision, Video, and Image Generation & SurveyX & 0.204 & 0.443 & 0.667 & 0.827 & 0.923 \\
    \hline
  \end{tabular}
  }
  \caption{\label{tab:recency_by_model_and_subtopic} Recency ratio comparison by model and subtopic.}
\end{table*}

The impact of different subtopics on survey metrics is mainly reflected in timeliness.
Table~\ref{tab:recency_by_model_and_subtopic} presents updated recency ratio comparisons across six major subtopics. 
\textit{SurveyGen-I} maintains consistently strong performance, ranking first in \texttt{RR@1} across all subtopics, and achieving top-2 results in nearly every other metric.
This indicates superior prioritization of the most recent literature compared to all baselines.
In particular, our model leads by a large margin on early recency in complex domains such as \textit{AI Applications}, \textit{Vision}, and \textit{Networks}.
While AutoSurvey occasionally matches or slightly outperforms on higher-k metrics (e.g., RR@10 in \textit{Databases}), its RR@1 remains substantially lower overall. 
However, this is partly due to the smaller number of available source papers in that subtopic, which reduces retrieval diversity and makes the evaluation more sensitive to a few recent citations. 
SurveyForge and SurveyX trail significantly, especially in foundational domains like \textit{Learning Algorithms} and \textit{Big Data}, reflecting weaker temporal grounding.

\section{Discussion on Metrics and Usage}

\paragraph{Note on Metric Usage and Purpose.}
Our evaluation incorporates several citation-related quantitative metrics, such as recency ratio, and citation density, to benchmark the structural and referential behavior of generated surveys. 
However, these indicators offer only partial signals of quality. 
Academic survey writing is inherently diverse and context-dependent: field maturity, topic breadth, and venue-specific formatting constraints (e.g., strict page limits) all shape citation practices and content scope. 
Moreover, certain high-quality surveys may intentionally limit citations to emphasize synthesis or conceptual framing. 
Thus, these tools collectively provide an empirical lens into model behavior, but should not be interpreted as exhaustive or definitive measures of writing quality.

\paragraph{Disclaimer on Intended Use.}
This tool is developed to assist researchers in efficiently exploring relevant literature and identifying topic structures. 
The generated content is not intended for direct use in scholarly publication. 
We cannot guarantee the factual correctness or citation fidelity of all outputs. 
This limitation is a known challenge in current LLM-based generation systems, especially for tasks involving factual grounding or citation synthesis.
Users remain responsible for verifying accuracy, attribution, and appropriateness. 
We explicitly discourage the direct submission of model-generated text for academic writing or peer-reviewed publication. All outputs require human review, editing, and contextual judgment.

\section{Examples}

\begin{figure}[th]
  \centering
  \includegraphics[width=\columnwidth]{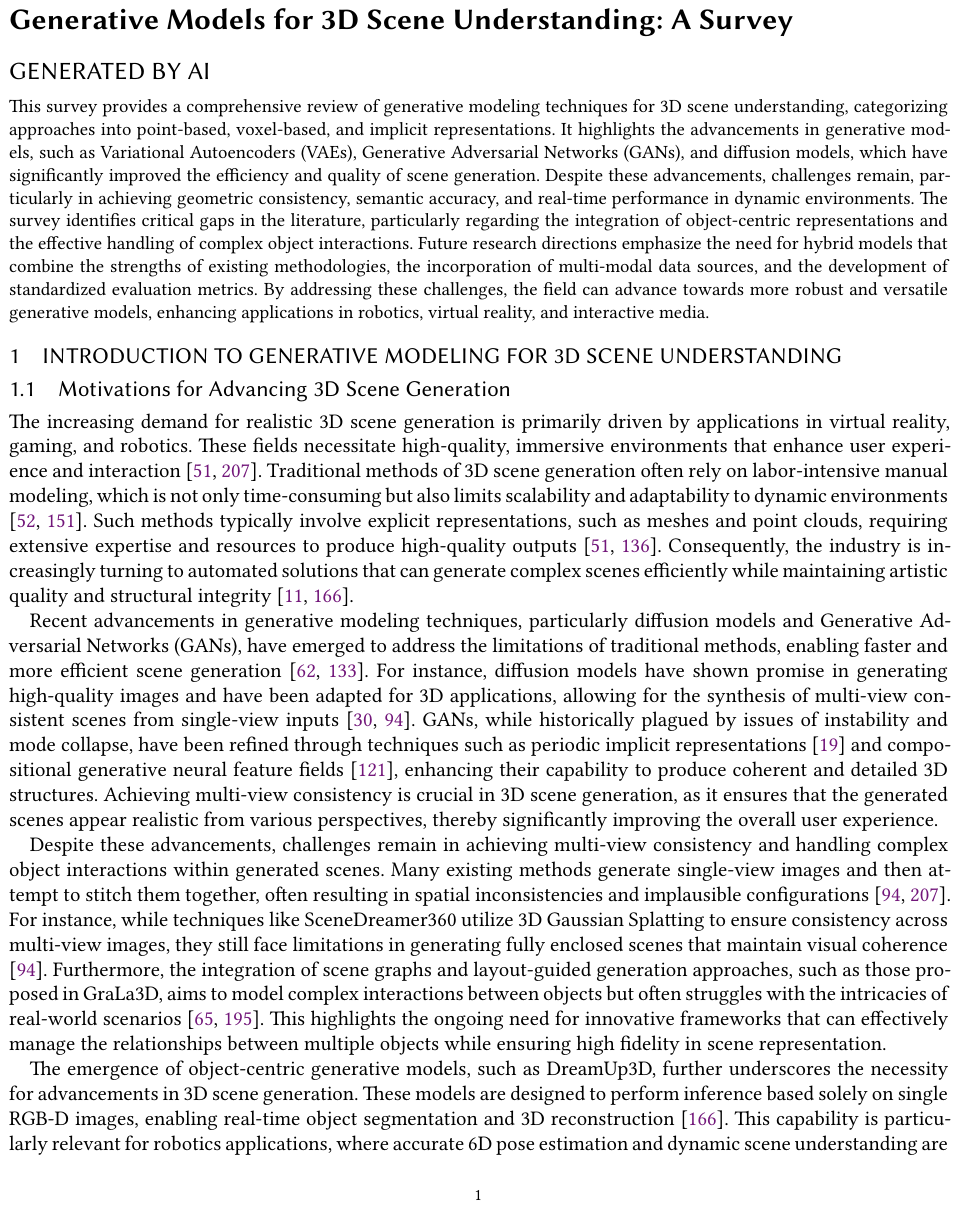}
  \caption{Front page generated by SurveyGen-I, demonstrating automatic formatting of title, author, and metadata.}
  \label{fig:cover}
\end{figure}

\begin{figure}[th]
  \centering
  \includegraphics[width=\columnwidth]{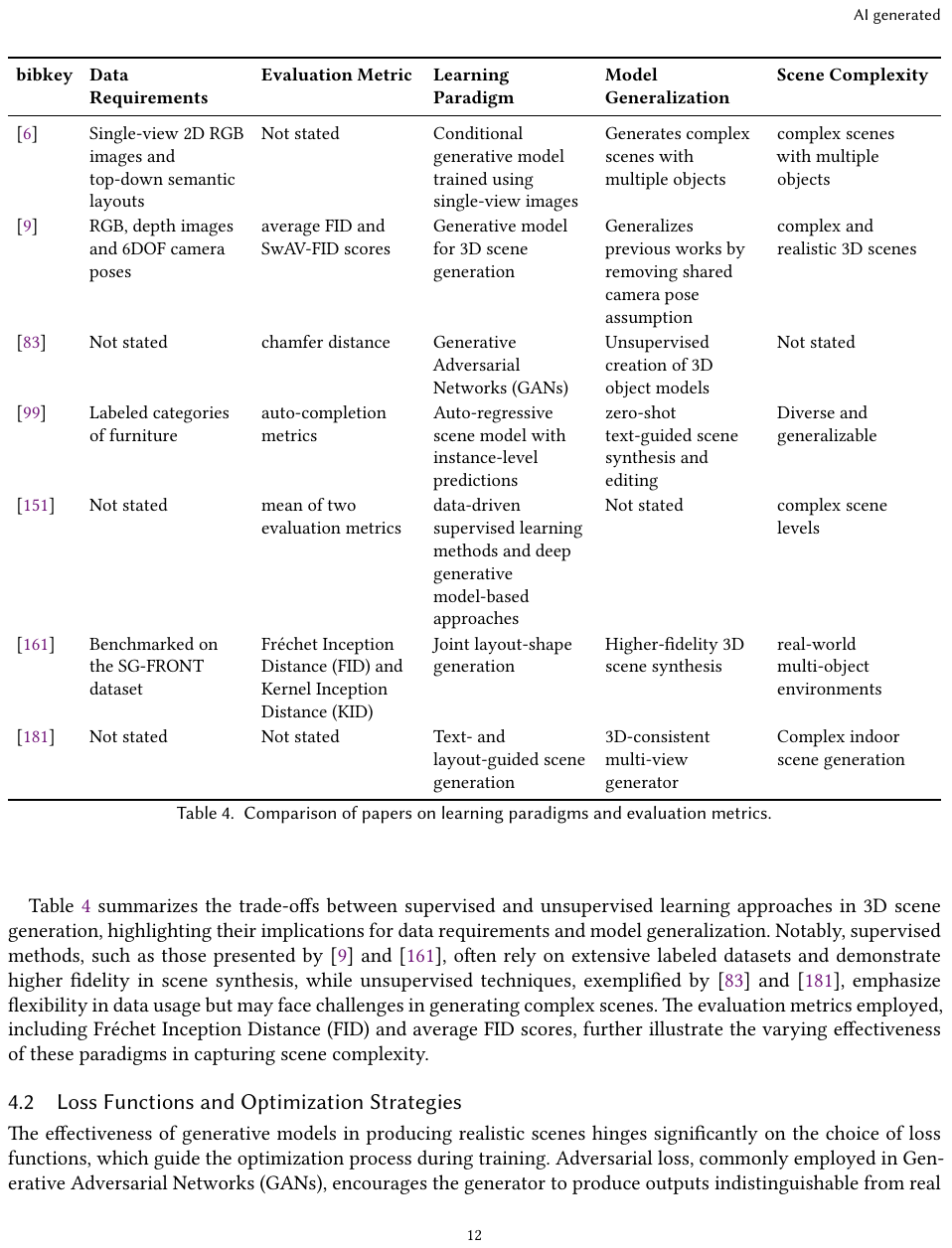}
  \caption{Citation alignment table produced by our system, showing mapped references and their summarized claims.}
  \label{fig:big_table}
\end{figure}

\begin{figure}[th]
  \centering
  \includegraphics[width=\columnwidth]{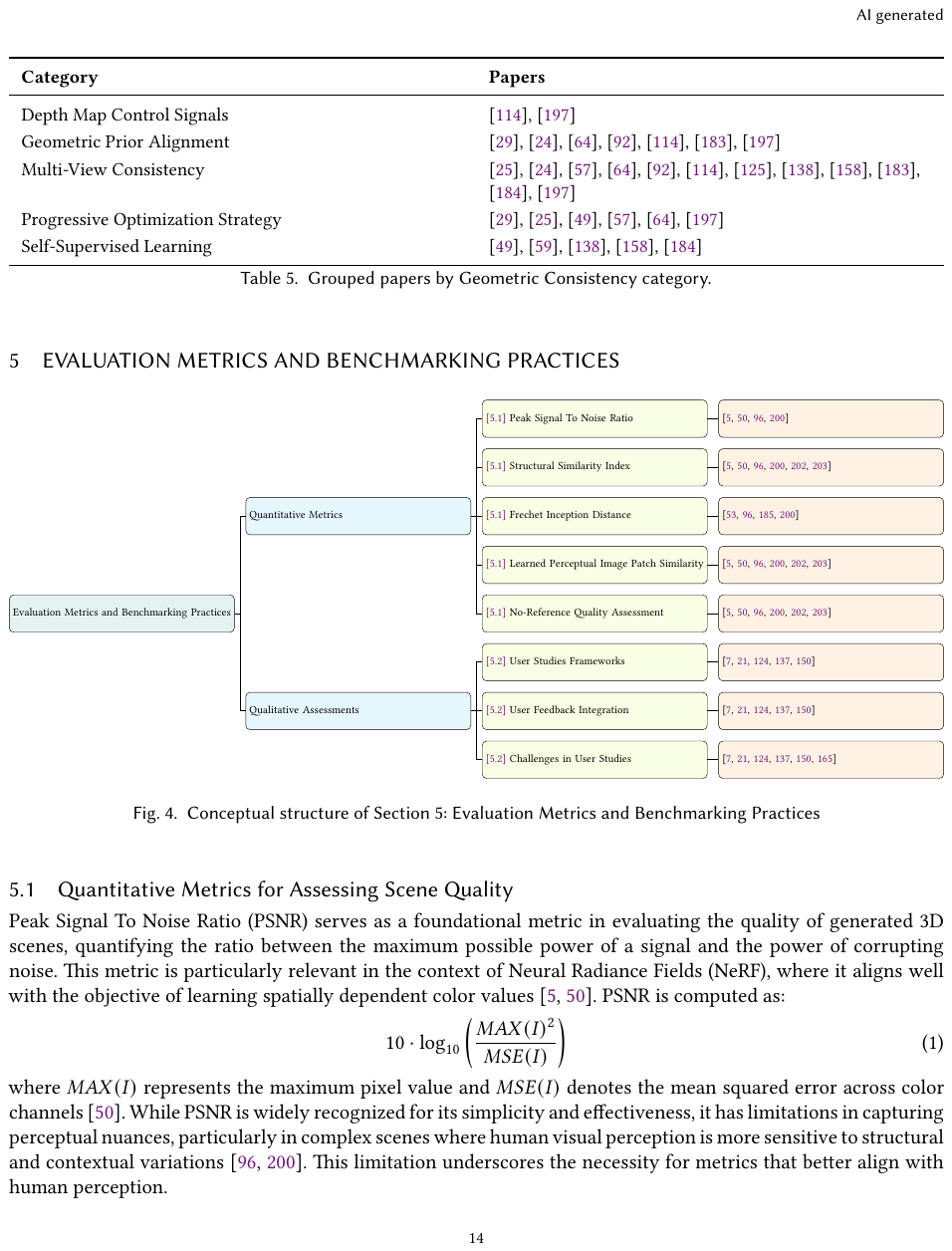}
  \caption{Example of complex formatting support, including tables, figures, and equations generated within a single subsection.}
  \label{fig:table_eq}
\end{figure}

\begin{figure}[th]
  \centering
  \includegraphics[width=\columnwidth]{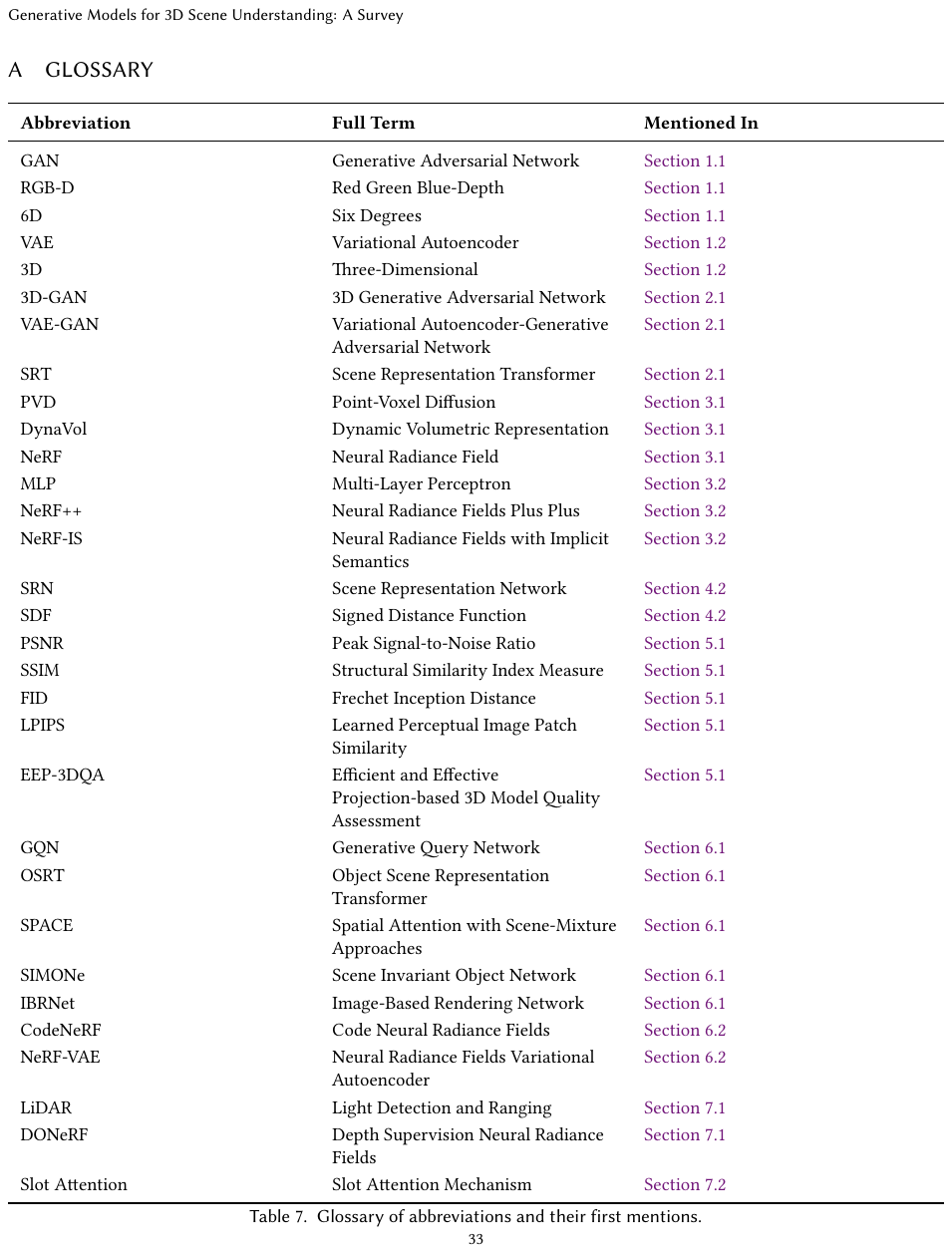}
  \caption{Glossary section automatically synthesized by SurveyGen-I to define key terms and acronyms.}
  \label{fig:glossary}
\end{figure}

\begin{figure*}[th]
  \centering
  \includegraphics[width=\textwidth]{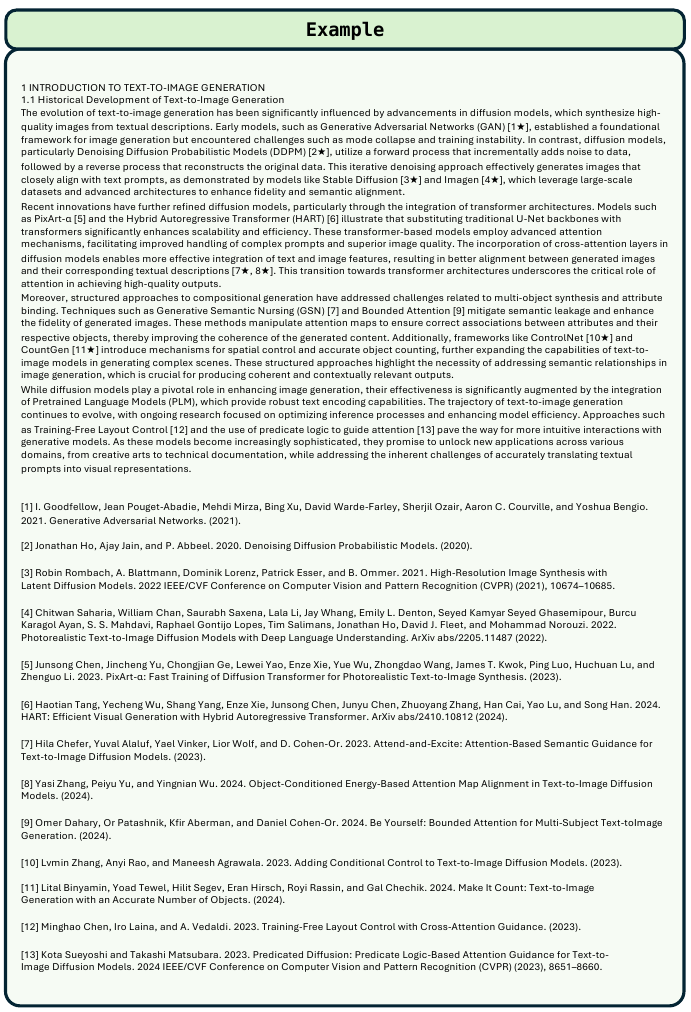}
  \caption{An example of generated example with SurveyGen-I. References marked with an asterisk (*) indicate citations that have been explicitly traced and verified. }
  \label{fig:example_paper}
\end{figure*}

Figures~\ref{fig:cover}, \ref{fig:big_table}, \ref{fig:table_eq}, and \ref{fig:glossary} illustrate different aspects of the generated surveys, including frontmatter design, citation table integration, mathematical expression formatting, and glossary usage. And Figure~\ref{fig:example_paper} presents an example output generated by our framework.

\section{Prompts}
\label{appendix:prompts}

\begin{figure}[th]
  \centering
  \includegraphics[width=\columnwidth]{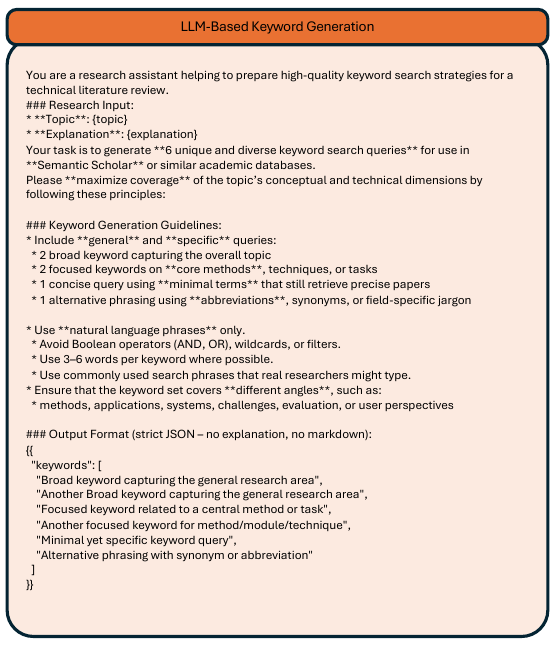}
  \caption{Keyword Generation used in SurveyGen-I.}
  \label{fig:key_word_generation_prompt}
\end{figure}

\begin{figure}[th]
  \centering
  \includegraphics[width=\columnwidth]{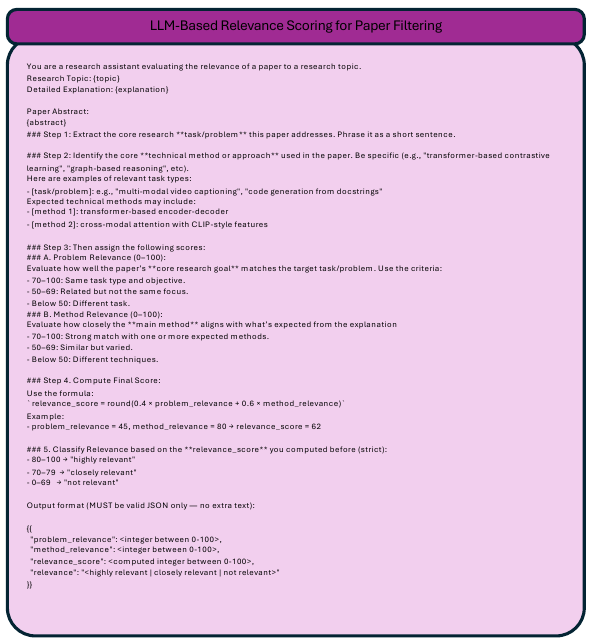}
  \caption{Relevance Scoring for Paper Filtering prompt used in SurveyGen-I.}
  \label{fig:relevance_scoring_for_filtering_prompt}
\end{figure}

\begin{figure}[th]
  \centering
  \includegraphics[width=\columnwidth]{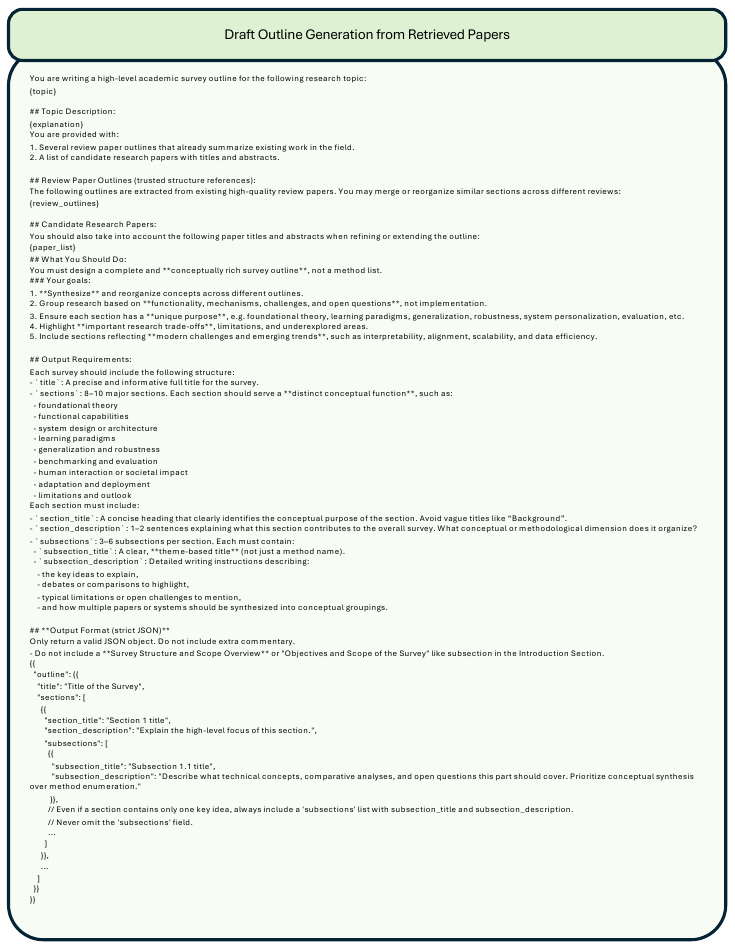}
  \caption{Draft Outline Generation from Retrieved Papers prompt used in SurveyGen-I.}
  \label{fig:relevance_scoring_prompt}
\end{figure}

\begin{figure}[th]
  \centering
  \includegraphics[width=\columnwidth]{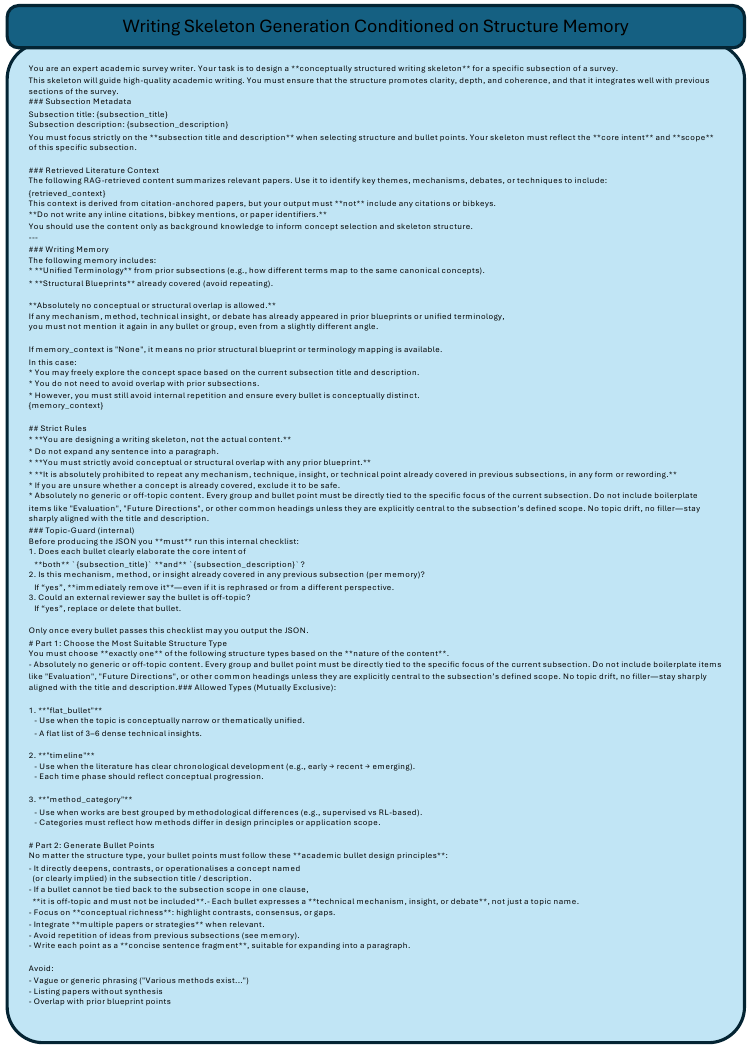}
  \caption{Writing Skeleton Generation Conditioned on Structure Memory prompt used in SurveyGen-I.}
  \label{fig:wirting_skeleton_generation_conditioned_on_structure_memory_prompt}
\end{figure}

\begin{figure}[th]
  \centering
  \includegraphics[width=\columnwidth]{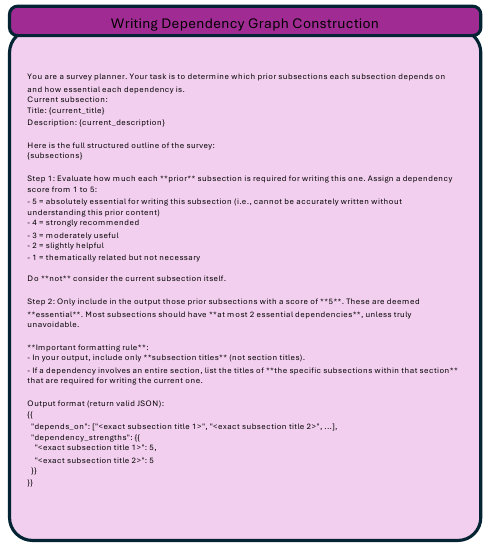}
  \caption{Writing Dependency Graph Construction prompt used in SurveyGen-I.}
  \label{fig:writing_dependency_graph_construction_prompt}
\end{figure}

\begin{figure}[th]
  \centering
  \includegraphics[width=\columnwidth]{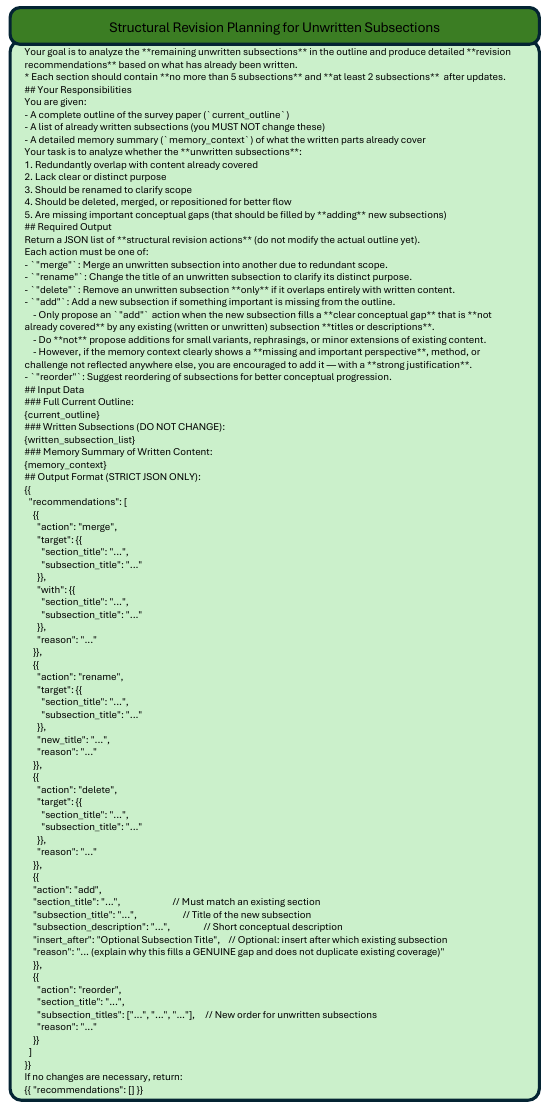}
  \caption{Structural Revision Planning for Unwritten Subsections prompt used in SurveyGen-I.}
  \label{fig:structural_revision_planning_for_unwritten_subsections_prompt}
\end{figure}

\begin{figure*}[th]
  \centering
  \includegraphics[width=\textwidth]{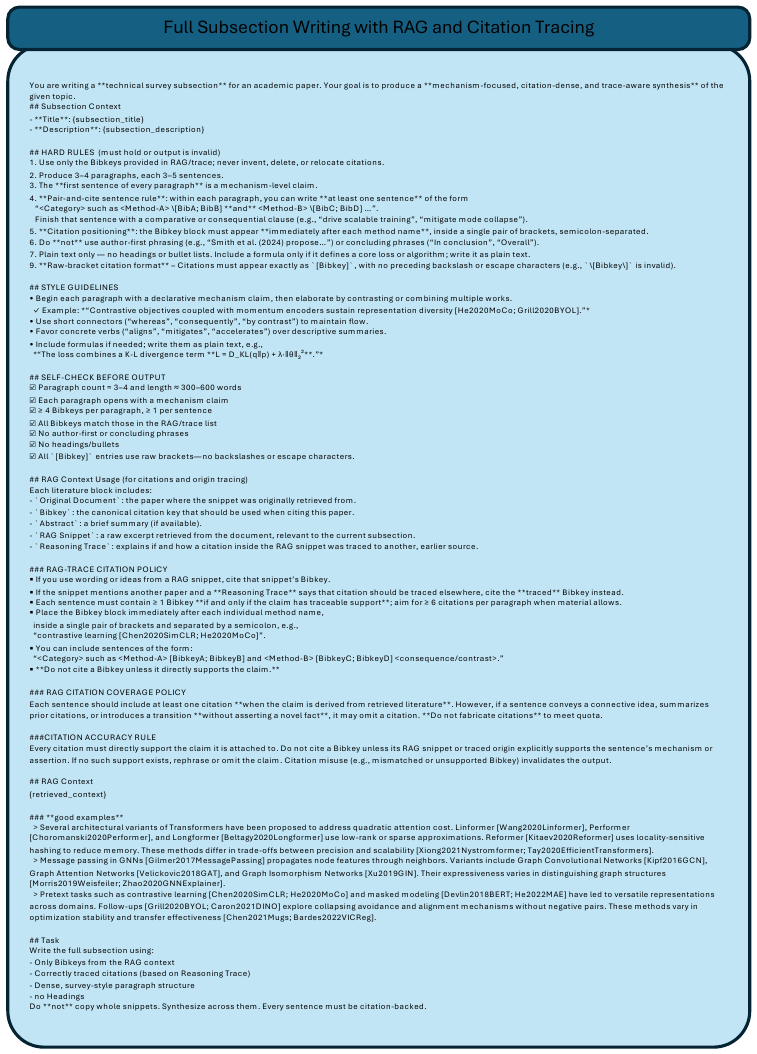}
  \caption{Full Subsection Writing with RAG and Citation Tracing prompt used in SurveyGen-I.}
  \label{fig:full_subsection_writing_with_rag_and_citation_tracing_prompt}
\end{figure*}

\begin{figure*}[th]
  \centering
  \includegraphics[width=\textwidth]{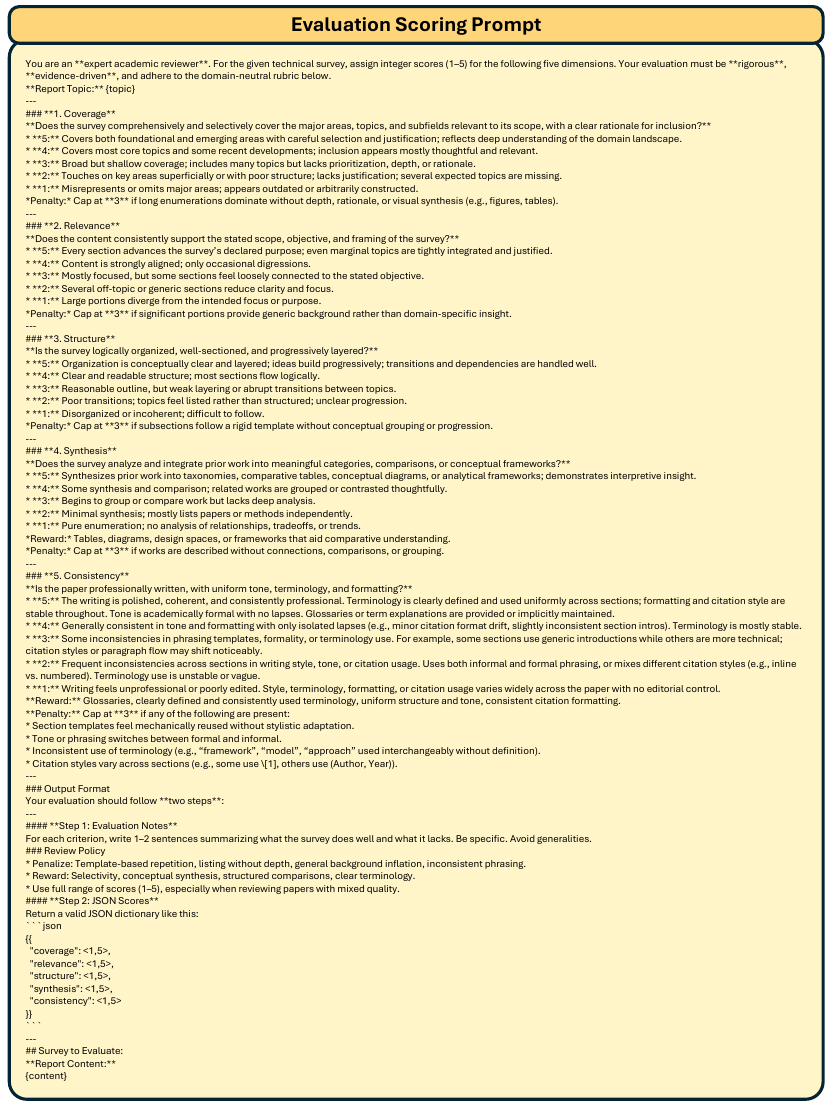}
  \caption{Evaluation Scoring prompt used in SurveyGen-I.}
  \label{fig:evaluation_scoring_prompt}
\end{figure*}

Figure~\ref{fig:key_word_generation_prompt}-\ref{fig:evaluation_scoring_prompt} show representative prompt designs in \textit{SurveyGen-I}; the full prompt suite is provided in the codebase.

\end{document}